\documentclass[twoside,11pt]{article}

\usepackage{dnd}
\usepackage{times}
\usepackage{graphicx}
\usepackage{times}
\usepackage{latexsym}
\usepackage{url}
\usepackage{amsmath,amssymb}
\usepackage{mathtools}

\usepackage{mathptmx}
\usepackage{breakcites}
\usepackage{amsmath}
\usepackage{color}
\usepackage{pdflscape}
\usepackage{longtable}
\usepackage{natbib}
\usepackage{booktabs}

\usepackage[colorinlistoftodos]{todonotes}
\usepackage{graphicx}
\usepackage{footnote}

\usepackage{enumitem,kantlipsum}

\dndheading{issue(number)}{year}{firstpage--lastpage}{Name1 Surname1, Name2 Surname2, and Name3 Surname3}{10.5087/dad.DOINUMBER}

\ShortHeadings{Looking to the Future of Natural Language Generation}{{Santhanam}, Sashank and {Shaikh}, Samira}
\firstpageno{1}

\begin{document}

\title{A Survey of Natural Language Generation Techniques with a Focus on Dialogue Systems - Past, Present and Future Directions}

\author{\name Sashank Santhanam \email ssantha1@uncc.edu \\
       \addr Department of Computer Science\\
       University of North Carolina at Charlotte
       \AND
       \name Samira Shaikh \email samirashaikh@uncc.edu \\
       \addr Department of Computer Science\\
       University of North Carolina at Charlotte}

\editor{Name Surname}
\submitted{MM/YYYY}{MM/YYYY}{MM/YYYY}

\maketitle

\begin{abstract}%
One of the hardest problems in the area of Natural Language Processing and Artificial Intelligence is automatically generating language that is coherent and understandable to humans. Teaching machines how to converse as humans do falls under the broad umbrella of Natural Language Generation.
Recent years have seen an unprecedented growth in the number of research articles published on this subject in conferences and journals both by academic and industry researchers. There have also been several workshops organized alongside top-tier NLP conferences dedicated specifically to this problem. All this activity makes it hard to clearly define the state of the field and reason about its future directions. 
In this work, we provide an overview of this important and thriving area, covering traditional approaches, statistical approaches and also approaches that use deep neural networks. 
We provide a comprehensive review towards building open domain dialogue systems, an important application of natural language generation. We find that, predominantly, the approaches for building dialogue systems use \textit{seq2seq} or \textit{language models} architecture. Notably, we identify three important areas of further research towards building more effective dialogue systems: 1) incorporating larger context, including conversation context and world knowledge; 2) adding personae or personality in the NLG system;  and 3) overcoming dull and generic responses that affect the quality of system-produced responses. We provide pointers on how to tackle these open problems through the use of cognitive architectures that mimic human language understanding and generation capabilities.  
\end{abstract}


\section{Introduction}
\label{intro}
\textit{Language Generation} is a sub-field of the field of Natural Language Processing (NLP), Artificial Intelligence (AI) and Cognitive Science (CS) that has been studied since the 1960s. NLG entails not only incorporating fundamental aspects of \textit{artificial intelligence} but also \textit{cognitive science}  \citep{reiter2000building}. Yet, it is still one of the major challenges towards achieving Artificial General Intelligence (AGI).  

\begin{figure}[h]
    \centering
    \includegraphics[height=8cm,width=10cm,keepaspectratio]{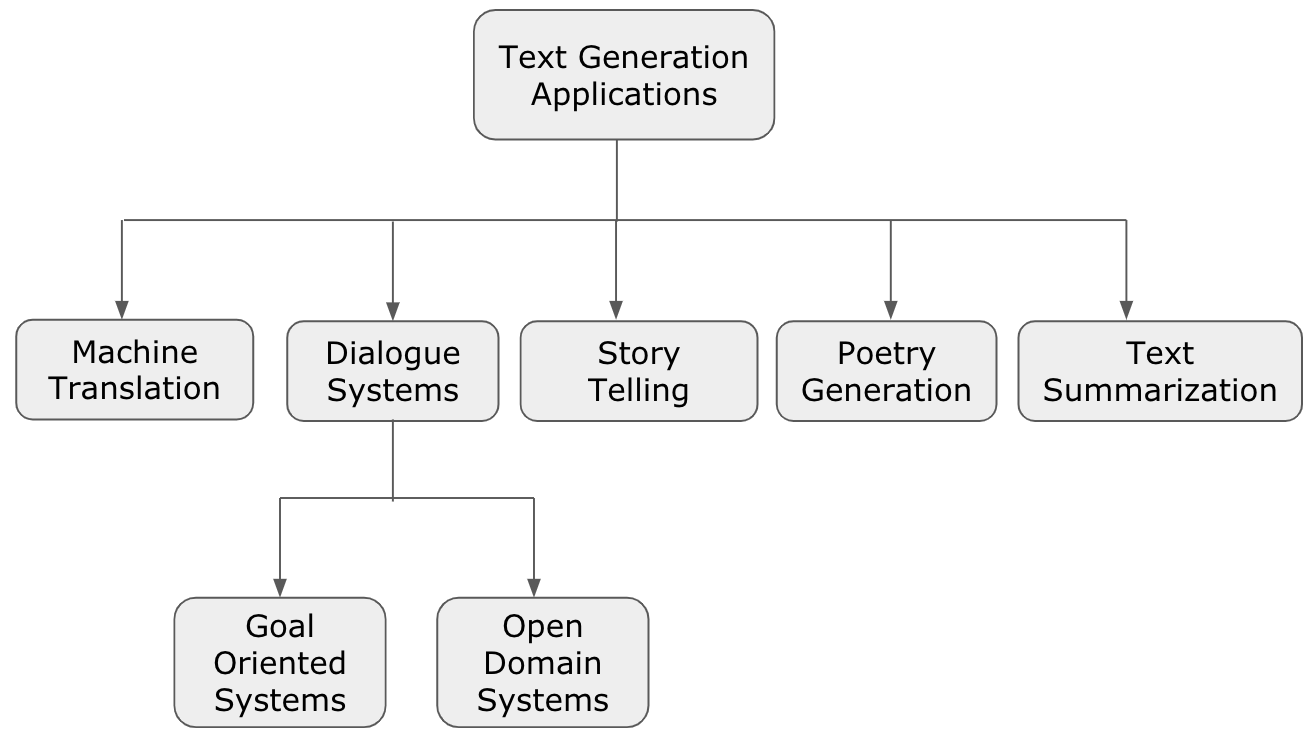}
    \caption{Categories in the sub-field of language generation.}  
    \label{fig:NLG_applications}
\end{figure}

In Figure \ref{fig:NLG_applications}, we provide an overview of the applications that can be categorized under the umbrella of language generation.  In this work, we focus on the domain of dialogue systems that is fundamental to Natural User Interfaces \citep{gao2019neural}.

Some of the early success in the field of language generation was building systems like Eliza \citep{weizenbaum1966eliza} and PARRY \citep{colby1975artificial}. These systems generated language through a set of rules. However, such rule based systems were too constrained and brittle and could not be easily generalized to produce diverse set of responses. Other traditional NLG techniques generated text from structured data or from knowledge bases. Some examples are domain-based systems that produce weather reports \citep{angeli2010simple} and sports reports \citep{barzilay2004catching}.

The field of text generation systems shifted from traditional approaches to statistical approaches where the focus was on exploiting patterns in text data and building models to make a prediction based on the text it has seen. Mikolov \emph{et al.} \citeyearpar{mikolov2010recurrent} argued that there had not been any significant progress in using statistical approaches to model language. This observation led to his experimentation on using recurrent neural networks \citep{mikolov2010recurrent} and achieved state-of-the-art results which set the wheels in motion for neural networks becoming a model of choice for modeling sequential data like text. Neural Networks belong to a class of machine learning models that are capable of identifying patterns in text and identify features that help solve different problems related to computer vision, object recognition, image captioning and speech recognition \citep{sutskever2014sequence}. Another phenomenon that suited the rise of neural networks is the large amount of corpora and significant computational resources that became available. In the applications of language generation, neural networks have helped achieve state-of-the-art results in problems related to machine translation \citep{bahdanau2014neural}, story telling \cite{holtzman2018learning}, dialogue systems \citep{wolf2019transfertransfo,xing2017topic,dinan2018wizard} and poetry generation \citep{zhang2014chinese}.

However, even with the powerful performance of neural networks for developing dialogue systems, current systems still suffer from problems like dull and generic responses \citep{li2015diversity}, lack of encoding context \citep{serban2016building,sordoni2015neural} and lack of consistent persona \citep{li2016persona}. Most current dialogue systems and conversational models lack style, which can be an issue as users may not be entirely satisfied with the interaction. Generating personalized dialogues is another substantially difficult task as the generated response has to be contextually-relevant to the conversation, while also conveying accurate paralinguistic features \citep{niu2018polite}. 

To make clear the directions towards which the field is heading, we produce a comprehensive overview of the field of open domain dialogue systems. Our primary goal is to identify the research gaps in the field and identify clear avenues for future research. While a few recent survey papers on this topic exist \citep{gatt2017survey,gao2019neural}, these do not identify the clear research gaps and also do not provide a comprehensive review of the field of open domain dialogue systems. 

In summary, the purpose of this paper is to: a) provide an overview of the research in the field of natural language generation from traditional approaches to deep learning based approaches (which we cover in Section~\ref{traditional-approaches} and Section~\ref{DL}); b) to give a comprehensive overview of the field of open domain dialogue systems (which we summarize in Table \ref{Survey} in Section~\ref{DS},);  and c) to propose avenues for future research for tackling these open problems (in Section \ref{sumftd}). 


\section{Traditional Approaches to Language Generation}
\label{traditional-approaches}
\begin{figure}[h]
    \centering
    \includegraphics[height=10cm,width=12cm,keepaspectratio]{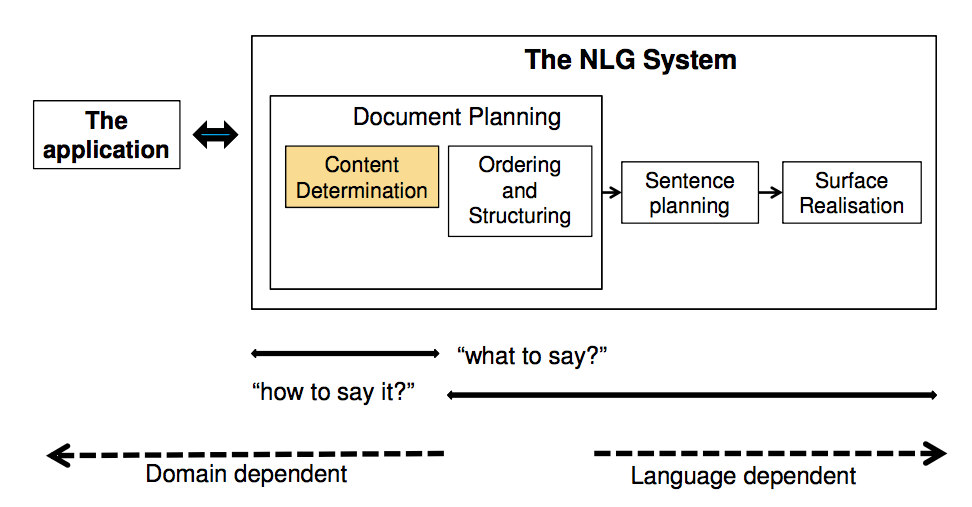}
    \caption{Three stage pipeline architecture proposed by Reiter and Dale focusing on the aspects of document planning, sentence planning and surface realization.} 
    \label{fig:NLG_arch}
\end{figure}

Reiter and Dale \citeyearpar{reiter2000building}  defined Natural Language Generation (NLG) as \textit{``the sub-field of artificial intelligence and computational linguistics that is concerned with the construction of computer systems than can produce understandable texts in English or other human languages from some underlying non-linguistic representation of information''}. They also presented a standard architecture for the developing NLG systems (Figure \ref{fig:NLG_arch}\footnote{\url{https://tinyurl.com/ydgyawvw}}) that comprised of six components each performing an important task to generate a coherent output. 

Their architecture was motivated by the fact that there were many NLG systems at the time for different applications but no well-defined, comprehensive architecture. Before the six stage pipeline, Reiter introduced a simple three stage pipeline of 1) content determination; 2) sentence planning and 3) surface realisation and named it the ``consensus'' architecture \citep{reiter1994has}. Cahill \emph{et al.} (\citeyear{cahill1999search}) conducted experiments and argued that the pipeline process was not detailed and the architecture was too constrained. In order to  overcome the issues, the authors suggested a finer architecture based on the linguistic operations such as 1) lexicalisation; 2) referring expression generation and 3) aggregation \citep{cahill1999search}. One drawback of the architecture suggested by Cahill \emph{et al.} \citeyearpar{cahill1999search} was that no details were provided about how the systems get input and in what form. Reiter and Dale \citep{reiter2000building} iterated on their initial architecture and suggested a new standard architecture for the NLG systems comprising of 4-tuple $\langle k,c,u,d\rangle$ where $k$ is the knowledge source, $c$ is the communicative model, $u$ is the user model and $d$ is the discourse theory \citep{evans2002nlg} and the iterated model also implemented some of the aspects of Cahill \emph{et al.} \citeyearpar{cahill1999search} work into the architecture.

In the following sub-sections, we explain the functionality of the six components and extensive research work that has been carried out to address that component.

\subsection{Content Determination}
Content Determination is the problem of deciding the domain that is needed to generate text for a given input. Content determination is affected by communicative goals i.e., different communicative goals from different kinds of people may require different contents to be expressed by the system that satisfy the parties involved.
Content determination is affected by the expertise of the end user and also by the content of the information source present in the system \citep{reiter2000building}. 

The problem of content determination has been approached from two 
different perspectives: 1) Schemas or templates; and 2) Statistical data driven approaches. 

\textbf{Schema- or Template-based content determination} methods focused on generating content by an analysis of the corpora and they are prominent in tasks while are standardised like weather forecast systems like FOG \citep{goldberg1994using} where rhetorical relations can be encoded as schemas or schemata \citep{mckeown1985discourse}. The schemata is made up of \textit{identification, constituency, attributive and contrastive}. Each component of the schemata is used to describe different predicate patterns \citep{mckeown1985discourse}. Schemas or Templates can be improved upon by using rule based approaches. Rule based approaches for the task of content determination have been used for domain specific systems where the implicit knowledge of the domain expert is used for more knowledge acquisition \citep{reiter2000knowledge,elhadad1996overview}.  Reiter \emph{et al.} \citeyearpar{reiter2000knowledge} list the different techniques such as sorting, thinking aloud, expert revision for knowledge acquisition in the STOP system that generated personalized smoking-cessation leaflets. 

With the availability of more data, the process of \textbf{content determination became data-driven}. 
Duboue and McKeown \citeyearpar{duboue2003statistical} developed a system that automated the process of producing constraints on every input and deciding if it should appear as a part of the output with the help of a two-stage process of exact matching and statistical selection, where the semantic data is clustered and text corresponding to each cluster is used to measure its degree of influence with regards to the other clusters. An alternative method was suggested by Barzilay and Lee  \citeyearpar{barzilay2004catching}, so that content selection can be applied to domains where the knowledge base has not been provided by using a novel adaptation of Hidden Markov Models. In their method, the states of the Hidden Markov models correspond to the type of information characteristic to the domain of interest. Barzilay and Lapata  \citeyearpar{barzilay2005collective} suggested another method along similar lines, in which the content selection is treated as a collective classification problem by capturing the contextual dependencies between the input items.
    
Liang \emph{et al.} (\citeyear{liang2009learning}) extended the work done by Barzilay and Lapata, by describing a probabilistic generative model that combines text segmentation and fact identification in a single unified framework using Hidden Markov Models. They proposed a generative model consisting of three stages of selecting a set of records, identifying the fields from the records and choosing a sequence of words from the fields and each stage, optimized using Expectation Maximization (EM). The work done by Liang \emph{et al.} \citeyearpar{liang2009learning} proved instrumental in combining the process of content determination and linguistic realization into a unified framework. Another example is the work done by Angeli \emph{et al.} \citeyearpar{angeli2010simple}, where the process of generation is broken down into a sequence of local decisions and using a classifier on decisions that include choosing records from the database, choosing a subset of fields from records and choosing a template to render the generated text . However, Kim and Mooney \citeyearpar{kim2010generative} identified a drawback with the method suggested by Liang \emph{et al.} \citeyearpar{liang2009learning} of just using bag of words and a simple Hidden Markov Model and not considering the context-free linguistic syntax. To address this issue, Kim and Mooney used a generative model with hybrid trees which expresses correspondence between the word in natural language and grammatical structure (meaning representation) and iterative generation strategy learning (a method similar to EM that iteratively improves probability to determine which event likely to be received as input from the human). Another example of content determination (in an end-to-end system) is the work done by Konstas and Lapata  \citeyearpar{konstas2012unsupervised}, where a set of records are converted into probabilistic context free grammar that describes the structure of the database and the grammar is encoded as a weighted hypergraph. The generation process is based upon finding the best derivation of the hypergraph. In the next section, we will cover document structuring, the next sub-problem of language generation.

\subsection{Document Structuring}
The second sub-problem specified by Reiter and Dale \citeyearpar{reiter2000building} is document or text structuring. This is the process of determining the order in which the text is to be conveyed back to the user once the content is determined. Document Structuring and Content Determination are closely linked. 


A method which had a significant impact on addressing this problem was the understanding of discourse relations with the help of Rhetorical Structure Theory (RST) \citep{mann1986relational}. RST has four elements consisting of ``relations'' which identifies relationships between different parts of the text in the form of satellite and nuclei. Nuclei represents the important part of the text and satellite represents the supplementary part of the text, ``schemas'' defines patterns in a part of text can be analyzed with regards to other spans (nodes of a tree), ``schemas application'' and ``structures'' and help in creating coherent texts \citep{mann1987rhetorical}.  

Moore and Paris \citeyearpar{moore1993planning} found issues with using RST when they tried to use the individual segments and rhetorical relations between segments to construct a text plan for their dialogue system. RST were not able to generate proper responses for follow up questions. Due to these problems with RST, Moore and Pollack \citeyearpar{moore1992problem} suggested a two-level discourse analysis process. The first level is called ``information level'' which involves the relation conveyed between two sentences in a discourse and second level is called ``intentional level'' which deals with the discourse produced to effect change in the mental state of the participants \citep{moore1992problem}. Dimitromanolaki and Androutsopoulos \citeyearpar{dimitromanolaki2003learning} used supervised machine learning to learn a new representation of document structuring task and applied this approach to for the task of document structuring for a specific domain.  
A lot of other researchers have interlinked the process of the text structuring and content determination into a single one which has been described in the previous subsection.

\subsection{Lexicalization}

Lexicalization or the task of choosing the right words to express the contents of the message is the third sub-problem defined by Reiter and Dale \citeyearpar{reiter2000building}. They broke down the task of lexicalization into two categories, namely, Conceptual Lexicalization and Expressive Lexicalization. Conceptual Lexicalization is defined as converting data into linguistically expressible concepts and Expressive Lexicalization is how lexemes available in a language can be used to represent a conceptual meaning \citep{reiter2000building}. In order to solve the problem choosing the best lexeme to realize the meaning, Bangalore and Rambow \citeyearpar{Bangalore:2000:CLC:1075218.1075277} suggested using a tree representation of the syntactic structure and an independently hand-crafted grammar. One of the drawbacks of this method was not using a part-of-speech tagger and using a mechanism of making a union of all the synonyms from the synset. 
While the traditional approaches to NLG view the process of lexicalization as belonging to the sentence planning phase along with the process of sentence aggregation and referring expression generation, however, recent research in NLG views lexicalization as the part of the linguistic realization phase \citep{gatt2017survey}.

\subsection{Referring Expression Generation}
 
Referring Expression Generation (REG) is the fourth sub-problem defined by Reiter and Dale \citeyearpar{reiter2000building} and it is aggregated with the sentence planning phase of the architecture. REG is the ability to produce a description of an entity and distinguish it from the other domain entities \citep{reiter2000building}. An entity might be referred to in many different ways. For example, consider the following sentence, {\footnotesize{\fontfamily{pcr}\selectfont{Adrian arrived late to an event and he missed a majority of it}}}. There can be two ways in which an entity can be referred to. The first is the \textit{initial reference}
({\footnotesize{\fontfamily{pcr}\selectfont{Adrian}}}) in the example) when the entity is brought into the discourse and the other is \textit{subsequent reference} ({\footnotesize{\fontfamily{pcr}\selectfont{he}}}) in the example) which refers to entity after it has been introduced in discourse \citep{reiter2000building}. The first step of the solution suggested by Reiter and Dale \citeyearpar{reiter2000building} is to identify the type of reference for the target, such as pronoun or description or proper name. The identification of proper names is the easiest, while identification of pronouns can be based on a rules such as ``the target is referred to in the previous sentence and if the sentence contained no other entity of the same gender'' \citep{krahmer2012computational}.


There are multiple existing algorithms for the task of REG. Dale \citeyearpar{dale1989cooking} created the \textbf{Full Brevity} algorithm that generates very short descriptions referring expression by the identification of target and distractors. However, this algorithm  suffered from major drawbacks such as being able to only generate short referring expressions and computing these short expressions had a high complexity (NP-hard)\citep{krahmer2012computational}. An improvement over the Full Brevity was the \textbf{Greedy Heuristic} algorithm, which picks a property of target that rules out most of the distractors (words that do not co-reference with the target) and adding that property to the description \citep{dale1992generating}. Greedy algorithm was later eclipsed in terms of performance by the \textbf{Incremental Algorithm} (IA). The Incremental Algorithm sequentially picks the properties and then rules out the distractors until a distinguishable expression is generated  \citep{dale1995computational}. However, these description generated may contain redundant properties which becomes a drawback of incremental algorithm.

To address these drawbacks, Kees and Van Deemter \citeyearpar{van2002generating} explored how the incompleteness of IA could be overcome with the help of a two stage algorithm to generate boolean descriptions. The first stage is the process of generalization of the IA by taking a union of the properties that help in singling out the target set and the next stage was to optimize the expressions produced \citep{van2002generating, krahmer2012computational}. One of the issues this work failed to address the notion of vagueness which was addressed in the work done by Horacek \citeyearpar{horacek2005generating}. Horacek \citeyearpar{horacek2005generating} introduced measures including the following to represent the uncertainties: $p_k$ - the user is acquainted with the terms mentioned, $p_p$- the user can perceive the properties uttered, $p_A$ - the user agrees with the applicability of the terms used. With the help of these three probabilities, the probability of recognition \textit{p} is calculated as the product of the three probabilities and this helps in distinguishing vagueness along with misinterpretation and ambiguity \citep{horacek2005generating}. Later, Khan \emph{et al.} \citeyearpar{khan2008generation} addressed the issue of structural ambiguity in coordinated phrases in the form of ``Adjective Noun1 Noun1'' to determine if the Adjective was associated with Noun1 or Noun2. Khan \emph{et al.} \citeyearpar{khan2008generation} conducted user studies and suggested how the generator can avoid these issues. However, Engonopoulos and Koller \citeyearpar{engonopoulos2014generating} argued that the listeners might misunderstand the generated expression. To address these concerns, Engonopoulos and Koller \citeyearpar{engonopoulos2014generating} proposed an algorithm to maximize the likelihood that a referring expression is understood by the user with the help of a probabilistic referring expression model $P(a \vert t)$, where t refers to the expression and a to the object in the domain. In the next subsection, we will cover the aspects of sentence aggregation which is dependent on the capability of REG algorithms.

\subsection{Sentence Aggregation}

Sentence aggregation is characterized as the process of removing redundant information during the generation of discourse without losing any information and to produce text in a concise, fluid and readable manner \citep{dalianis1999aggregation}. Dalianis, in his survey suggested that aggregation can be done in all the stages of the NLG process except during content determination and surface realization. Reiter and Dale marked this subproblem as belonging to the sentence planning or microplanning phase \citep{reiter2000building}. Reiter and Dale characterized the problem of aggregation to be closely interlinked with lexicalization as both deal with understanding the knowledge source and linguistic elements of words, phrases and sentences \citep{reiter2000building}. 

One of the initial approaches to tackle the problem of sentence aggregation was put forward by Cheng and Mellish \citeyearpar{cheng2000capturing} by using Genetic Algorithms, where they used a constraint-based program with a preference function to evaluate the coherence of a text. Walker \emph{et al.}. \citeyearpar{walker2001spot} used a data-driven approach to overcome the issue of using a hand-crafted preference function used by Cheng \emph{et al.} \citeyearpar{cheng2000capturing}. In their work, they used two phases; the first phase generated a large sample of sentences for an input and the next phase ranked the sentences with the help of rules generated from training data. Barzilay and Lapata  \citeyearpar{barzilay2006aggregation} presented an automatic method to learn the grouping constraints with the help of a parallel corpus of sentences and their corresponding database entries by looking at the number of attributes shared by the entries. In the next section, we cover the aspect of linguistic realization that is the final stage of the pipeline and the different mechanisms that operate on the work done win earlier stages of the pipeline.

\subsection{Linguistic Realization}
Linguistic Realization was characterized by Reiter and Dale as the task of ordering different parts of a sentence and using the right morphology along with punctuation marks which is governed by rules of grammar to produce a syntactically and orthographically correct text \citep{reiter2000building}. In this section, we will cover three approaches for linguistic realization.

\subsubsection{Hand-coded grammar-based systems}
Grammar-based NLG systems are systems that make their choice depending on the grammar of the language, which can be manually written with the help of multilingual realizers. An example of multilingual realizer is KPML, developed by Bateman \citep{gatt2017survey, bateman1997enabling}, that depended on the systemic grammar to help understand the syntactic characteristic of a sentence. Another popular realizer, SURGE, was developed by Elhadad and Robin \citeyearpar{elhadad1996overview}, based on functional unification formalism. Another popular realizer was called Halogen, which was introduced by Langkilde \citeyearpar{langkilde2002halogen}. This system uses a small set of hand-crafted grammar rules as features to generate alternative representations. A downside of using these realizers is that they are complicated to use and have a steep learning curve for the users, which made the NLG community move towards simple realization engines.

\subsubsection{Templates}

Templates are often used in systems which require limited syntactic variability in their output \citep{reiter1997building}. Consider the template {\footnotesize{\fontfamily{pcr}\selectfont{$[person]$ is leaving $[country]$}}} and in this scenario {\footnotesize{\fontfamily{pcr}\selectfont{person}}} and {\footnotesize{\fontfamily{pcr}\selectfont{country}}} values can be replaced by the system during output phase. One of the issues with template-based NLG systems is lack of flexibility of the templates to produce a diverse set of generated texts. McRoy \emph{et al.}  \citeyearpar{mcroy2003augmented} suggested a method to overcome these issues with the help of declarative control expressions to augment traditional templates. Van Deemter \emph{et al.} \citeyearpar{van2005real} argued that as new NLG systems have been developed, the differences between standard NLG systems and template-based systems have blurred as the modern systems use handcrafted grammars to help with realization. Another disadvantage of using templates is the need for knowledge expertise to construct templates for the system \citep{mcroy2003augmented,gatt2017survey}. Angeli \emph{et al.} \citeyearpar{angeli2012parsing} used a probabilistic approach and compositional grammar to learn the rules for parsing time expressions.  Kondadadi \emph{et al.} \citeyearpar{kondadadi2013statistical} used \textit{k}-means clustering to create template banks derived using named entity tagging and semantic analysis. Despite the advantage of using template based methods, most of the recent NLG systems have moved to a statistical-based approach.

\subsubsection{Statistical Approaches}
Statistical approaches have been used in NLG systems in order to reduce the manual effort of using hand written grammar rules and to deal with large corpora to acquire probabilistic grammar to get better realizations of text. The work by Langkilde \citeyearpar{langkilde2000forest} was one of the seminal works in using statistical approaches towards linguistic realization. In this approach, Langkilde \citeyearpar{langkilde2000forest} used corpus based statistical knowledge and a small hand crafted grammar to generate many different representations of a sentences that were packed in the form of forest of trees. Langkilde \citeyearpar{langkilde2000forest} ranked each phrase by calculating a score which was decomposed into a internal and external score, former known to be context independent and latter was context dependent. This method introduced by Langkilde served as the base for subsequent research in this field. 

Another important work was carried out by Langkilde and Knight \citeyearpar{langkilde1998generation}, to build a generator by computing word lattices from meaning representations by introducing new grammar formalisms. Bangalore and Rambow \citeyearpar{Bangalore:2000:CLC:1075218.1075277} suggested improvements by introducing a tree-based model of syntactic representation along with independently hand-crafted grammar rules to improve to performance of the syntactic choice module. Cahill \emph{et al.} \citeyearpar{cahill2007stochastic} presented a different method to rank and suggested using a log-linear ranking system, and they show that log-linear ranking obtained better performance than existing systems. 

One major downside of these approaches listed above is that the they are computationally expensive, as they generate a lot of possible sentence and then do the filtering with the help of the ranking mechanism. To overcome this drawback, Belz and Anja  \citeyearpar{belz2008automatic} introduced the Probabilistic Context-free Representationally Underspecified (pCRU) which uses probabilistic choice to inform generation instead of listing all the choices and then selecting a phrase.

The approaches described above all use a set of hand-crafted rules as the base generation and only use statistical method for the filtering the output. An alternative would be to apply statistical approaches on the base-generation systems. There have been approaches where grammatical rules have been derived from treebanks \citep{gatt2017survey}. Hockenmaier and Steedman \citeyearpar{hockenmaier2007ccgbank} presented a method to extract dependencies and combinatory categorical grammar(CCG) from the Penn Treebank corpus. 

Having given an overview of the traditional methods used for NLG and also the methods to address the subcomponents of the language generation process, in the next subsection we cover deep neural networks and the recent surge in these architectures towards solving natural language generation problem.

\section{Deep Learning approaches for Language Generation}
\label{DL}
Applying deep neural networks to Natural Language Processing has helped achieve state-of-the-art performance across different tasks, including the task of language generation due to the capability of neural networks to learn representations with different levels of abstraction \citep{lecun2015deep,goldberg2016primer}. The simplest and most widely used type of neural network is the feed forward neural network or multilayer perceptron \citep{rosenblatt1958perceptron} in which the data flow is in one direction and feed forward neural networks are acyclic graph structures. Bengio \emph{et al.} \citeyearpar{bengio2003neural} demonstrated the ability of feed forward neural networks on language modeling tasks. 
Another type of neural network architecture that is more suited for dealing with sequential data 
is the Recurrent Neural Network (RNN) architecture. RNNs are used for the processing of sequential data with the help of recurrent connections that perform the same task over every sequence \citep{Goodfellow-et-al-2016}. RNNs have the capability to handle long sequences using the knowledge gained (\textbf{``memory''}) from previous sequence computations unlike networks without sequence-based specialization. Application of memory to neural networks was demonstrated as early as 1982, through the Hopfield Network that was used to store and retrieve memory from a pre-trained set of patterns or memories, similar to the human brain. The network relied on neurons each producing a value of +1 or -1 depending on the input from the previous layer \citep{hopfield1982neural}. 

\begin{figure}[h]
    \centering
    \includegraphics[trim=0 80 0 0,clip,width=\columnwidth]{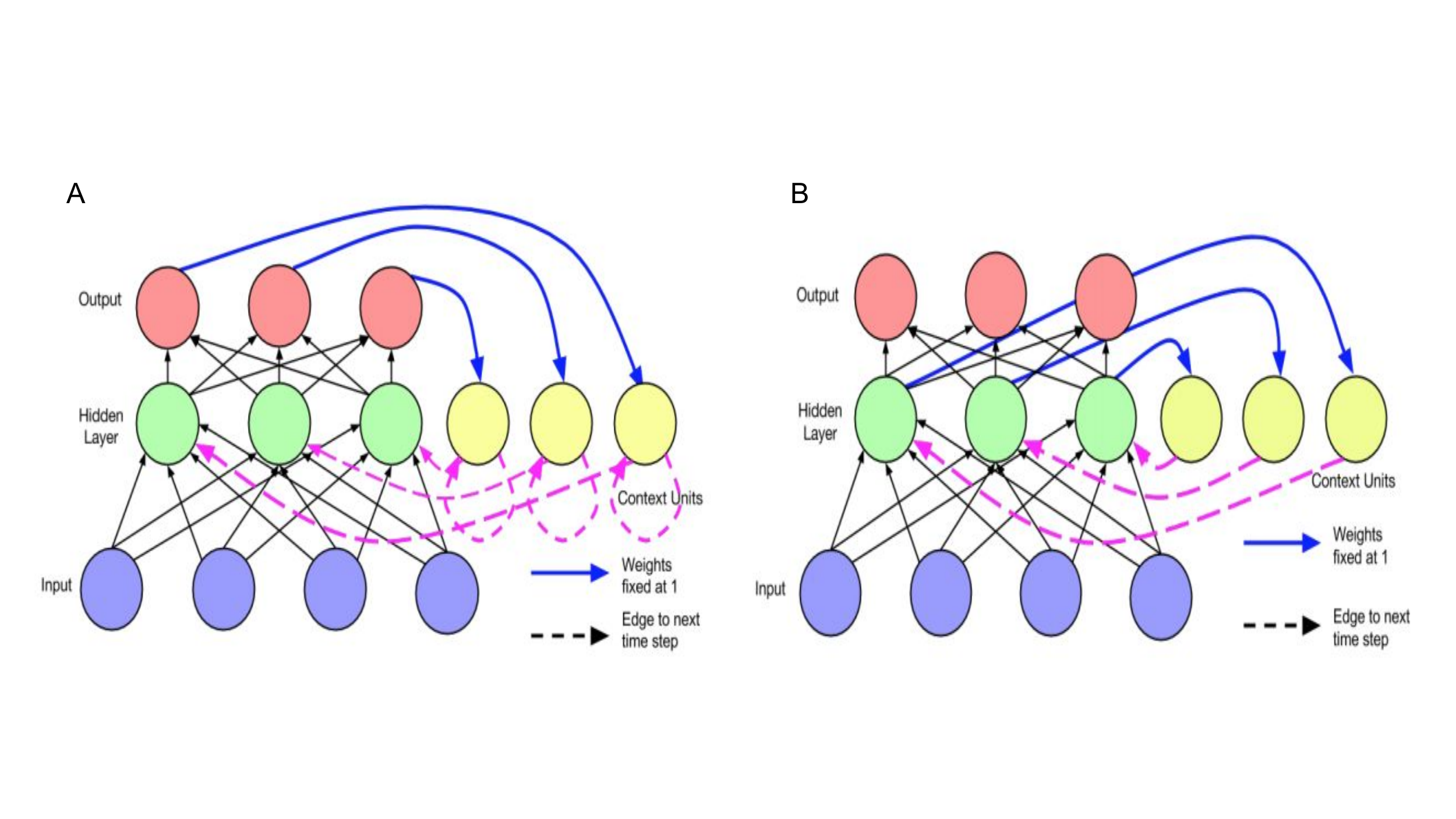}
    \caption{A. Represents the Jordan Architecture.
             B. Represents the Elman Architecture (Figure credit: Lipton \emph{et al.}, \citeyearpar{lipton2015critical})}
    \label{fig:JordanElman}
\end{figure}

Hopfield's network was the inspiration behind Jordan's network \citep{jordan1986serial} (represented in Figure \ref{fig:JordanElman}A), for doing supervised learning on sequences with the help of a single hidden layer and special units which receive input from the output unit which then forwards the values to the hidden nodes \citep{lipton2015critical}. Elman simplified the Jordan's architecture (represented in Figure \ref{fig:JordanElman}B), by adding a context unit with each hidden unit receiving its input from the units at the previous time step. Elman showed that network can learn dependencies by training the network on sequence of 3000 bits. The model achieved an accuracy rate of $66.7\%$ on predicting the third bit in the sequence \citep{elman1990finding, lipton2015critical}. 

The Elman architecture played a substantial role in the discovery of long short term memory networks (LSTM) \citep{hochreiter1997long}. LSTMs helped in tackling the important problems of vanishing and exploding gradients caused by backpropagation while training the neural networks \citep{schmidhuber2015deep}. During backpropagation, the neural network weights receive an update proportional to the gradient of the error function. These gradients are multiplied across layers and sometimes the gradients become too small or vanish and in certain cases the gradients grow exponentially and explode. LSTM replaced the hidden units of the neural networks with a new concept called memory cell, which is built around a central linear unit (internal state) with a fixed self connection to ensure that the gradients can pass without exploding or vanishing. The memory cell also contains an input and output gate; later the forget gate was added to the structure of the memory cell by Gers \emph{et al.} \citeyearpar{gers1999learning}. Gates are regulating structures that carefully allow the information to the internal state to be added or removed.

\begin{figure}[h]
    \centering
    \includegraphics[trim=0 0 0 20,clip,height=5cm,width=10cm]{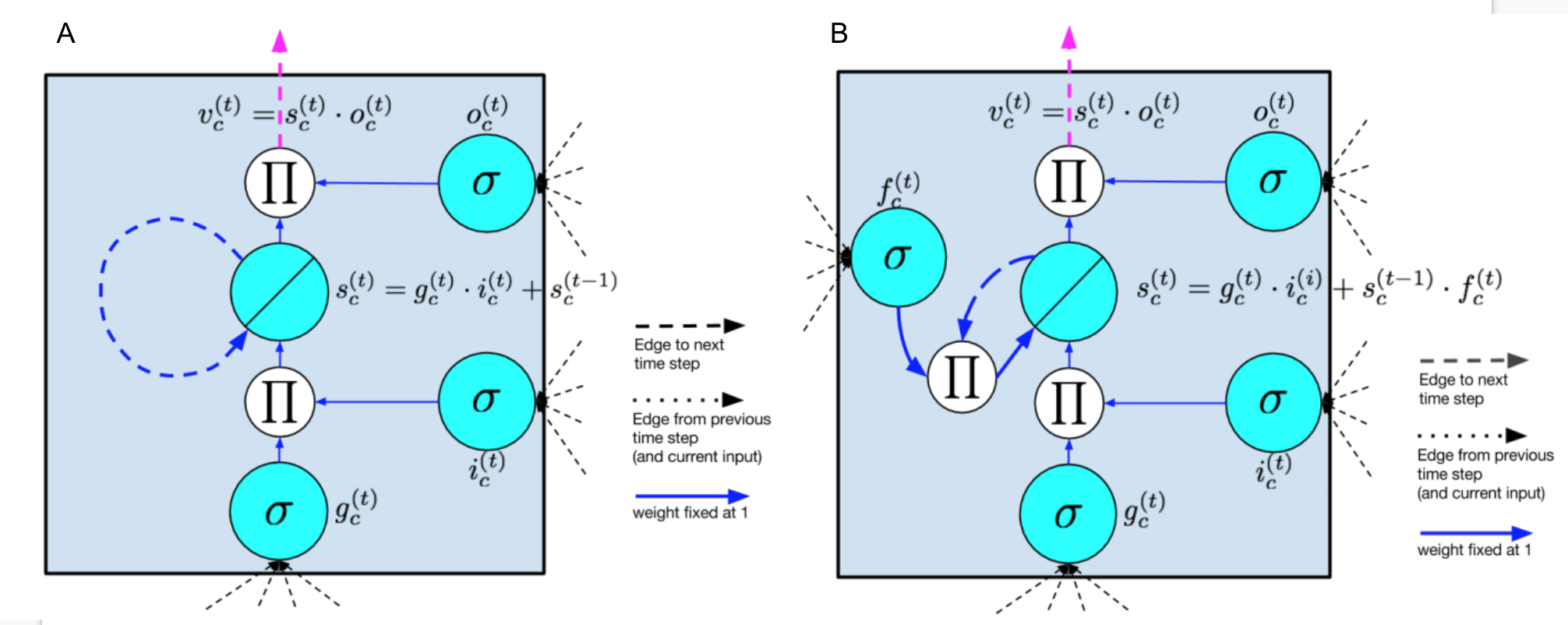}
    \caption{A. Represents the LSTM Architecture by Horcheiter \emph{et al.} \citeyearpar{hochreiter1997long}.
             B. Represents the updated LSTM Architecture by Gers \emph{et al.}\citeyearpar{gers1999learning}. (1) is the input node, (2) is the input gate, (3) is the output gate, (4) is the cell state, (5) is the forget gate. (Figure credit: Lipton \emph{et al.,} \citeyearpar{lipton2015critical}}
    \label{fig:LSTM}
\end{figure}
\newpage
We describe the various components shown in Figure~\ref{fig:LSTM} next:
\begin{itemize}
    \item Input node \textendash  The input nodes takes in the input from current layer $x^{(t)}$, t represents the current time step and also takes in the value from the hidden layer at the previous time step $h_{(t-1)}$ and the weighted sum input is taken is passed through an activation function ``sigmoid'', which was replaced by ``tanh'' as the LSTM architecture was improved.
    \begin{equation}
        g_c^{(t)} = \sigma (W_g.[x^{(t)},h_{(t-1)}] + b_g)
    \end{equation}
    \item Input gate \textendash  The input gate takes the input from the current layer $x^{(t)}$ and value from the hidden layer at the previous time step $h_{(t-1)}$ and applies a ``sigmoid'' activation function to the weighted sum. A sigmoid is used as the gate in this situation to make sure that any value that is a 0 then the corresponding value from the input gate is also cut off and cannot affect the internal state update.
    \begin{equation}
        i_c^{(t)} = \sigma (W_i.[x^{(t)},h_{(t-1)}] + b_i)
    \end{equation}
    \item Forget gate \textendash  The forget gate was added to the LSTM architecture to overcome a limitation of the the cell state growing linearly and when presented with a continuous stream the cell state might grow in an unbounded station \citep{gers1999learning}. The main job of the forget gate is to provide with a way to reset the contents of the cell state.
    \begin{equation}
        f_c^{(t)} = \sigma (W_f.[x^{(t)},h_{(t-1)}] + b_f)
    \end{equation}
    \item Cell State \textendash  The cell state is the heart of the memory cell and carries information that it has maintained until the current time step \textit{t} so that the loss function is not only dependent on the data from the current time step.
    \begin{equation}
        s_c^{(t)} = s_c^{(t-1)} \times f_c^{(t)} + g_c^{(t)} \times i_c^{(t)}
    \end{equation}
    \item Output Gate \textendash  The output gate takes the input from the current layer $x^{(t)}$ and value from the hidden layer at the previous time step $h_{(t-1)}$ and applies a sigmoid activation function to the weighted sum. A sigmoid is used as the gate in this situation to determine what values of the cell part of the cell state is to output.
    \begin{equation}
        o_c^{(t)} = \sigma (W_o.[x^{(t)},h_{(t-1)}] + b_o)
    \end{equation}
    \item Output Node \textendash The final output of the LSTM cell is obtained after passing the cell state through a tanh activation function and multiple it with the contents of the output gate.
    \begin{equation}
        v_c^{(t)} = o_c^{(t)} \times \tanh(s_c^{(t)})
    \end{equation}
\end{itemize}

\begin{figure}[h]
    \centering
    \includegraphics[width=5cm,height=4cm]{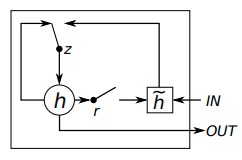}
    \caption{GRU architecture by Cho \emph{et al.}}
    \label{fig:GRU}
\end{figure}

Another variant of RNN model called \textbf{GRU} (See Fig \ref{fig:GRU} was introduced Cho \emph{et al.}\citeyearpar{cho2014learning}, inspired by the functionality of the LSTM. A performance comparison between the LSTM and GRU were conducted by Chung \emph{et al.} \citeyearpar{chung2014empirical} who found the performance between the two to be comparable. The GRU consists of two gates, reset (Eq \ref{reset_gru}) and update gates (Eq \ref{update_gru}) and exposes the whole state each time without having a mechanism to control it. The reset helps the hidden unit forget information not needed and the update gate controls how much information is carried forward from the previous hidden state. The actual activation is computed as a linear interpolation of the previous activation and candidate activation (Eq \ref{gru_activation}).

\begin{equation}
    r_{j} = \sigma([W_{r}.x]_{j} + [U_{r}.h_{(t-1)}]_{j}
    \label{reset_gru}
\end{equation}

\begin{equation}
    z_{j} = \sigma([W_{z}.x]_{j} + [U_{z}.h_{(t-1)}]_{j}
    \label{update_gru}
\end{equation}

\begin{equation}
    \begin{split}
        h_t^j = (1-z_t^j).h_{t-1}^j + z_t^j.\tilde{h}_t^j \\ 
        \tilde{h}_t^j = tanh(W.x_t + U(r_t \odot h_{t-1}))^j
    \end{split}
    \label{gru_activation}
\end{equation}

In the next four subsections, we list the different approaches such as language modeling, encoder-decoder, memory networks and transformer  models based approaches that have been applied to the task of language generation.

\subsection{Language Models}
Language models are probabilistic models that are capable of predicting the next word given the preceding words in a sequence. Language models are widely used in the generative modeling tasks. The ability of language models to model sequential data of fixed length context using feed forward neural networks was first demonstrated in the work done by Bengio \emph{et al.} \citeyearpar{bengio2003neural}. However, a major drawback of the approach, which is the usage of fixed length context, was overcome in the seminal work done by Mikolov \emph{et al.} \citeyearpar{mikolov2010recurrent} demonstrating the efficiency of RNN based language models. Similarly, another seminal work in the area of language models is the work done by Sutskever \emph{et al.}  \citeyearpar{sutskever2011generating} demonstrating the effectiveness of LSTM in predicting the next character of a sequence. Conditional language models are also used as a variant of language models where the language model is conditioned on variables other than the preceding words, like the work done by generating product reviews based on sentiment, author, item or category \citep{lipton2015generative} or generating text with emotional context \citep{ghosh2017affect}.  


\subsection{Encoder-Decoder Architecture}
\begin{figure}[h]
    \centering
    \includegraphics[height=6cm]{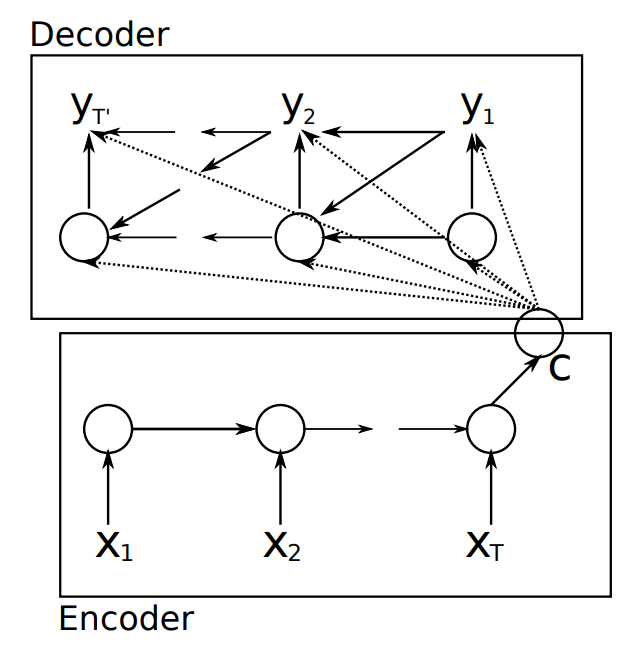}
    \caption{Encoder-Decoder architecture proposed by Cho \emph{et al.} (Figure credit: Cho \emph{et al.}\citeyearpar{DBLP:journals/corr/ChoMGBSB14}.)}
    \label{fig:enc}
\end{figure}

Another important architecture that enhanced the task of language generation was the usage of two RNNs in an end-to-end model (Figure \ref{fig:enc}) \citep{DBLP:journals/corr/ChoMGBSB14} that overcame a significant limitation where the neural networks could only be applied to problems where input and target can be encoded with fixed dimensionality. The \textbf{encoder} converts the input sequence into a fixed vector representation $c$ by Eq \ref{enc} where $h_t$ refers to hidden state at time step t, $f$ represents any non-linear function and $x$ represents the input sequence. The \textbf{decoder} tries to predict sequence of symbols with the help of the context vector $c$. The hidden state of the \textbf{decoder} depends on the context vector $c$ and is represented by Eq \ref{dec} and next symbol to be predicted is based on a condition probability \ref{enc-prob} where $g$ is a softmax function. 

\begin{equation}
    h_{(t)} = f(h_{(t-1)},x_t)
    \label{enc}
\end{equation}

\begin{equation}
    s_{i} = f(s_{i-1},y_{i-1},c)
    \label{dec}
\end{equation}

\begin{equation}
    P(y_t|y_{t-1},y_{t-2},...,y_1,c) = g(s_{(t)},y_{t-1},c)
    \label{enc-prob}
\end{equation}

Along similar lines to the work done by Cho \emph{et al.}   \citeyearpar{DBLP:journals/corr/ChoMGBSB14},  \textit{\textbf{seq2seq}} was introduced by Sutskever \emph{et al.}  \citeyearpar{sutskever2014sequence} which uses two LSTMs, one to map the input sequence to a fixed vector and the other RNN to decode the fixed vector into a sequence of target symbols of varying lengths. A key difference between the work done by Cho \emph{et al.} \citeyearpar{DBLP:journals/corr/ChoMGBSB14} and Sutskever \emph{et al.}   \citeyearpar{sutskever2014sequence} was the discovery that reversing the order of the input sequence improves the performance of the model and also helps with creating short term dependencies between input and target sequence. Bahdanau \emph{et al.}  \citeyearpar{bahdanau2014neural} identified the bottleneck caused by encoding the entire sequence into a fixed vector in the simple encoder-decoder architecture and proposed a modification which allows the decoder to \textit{attend} to different parts of the source sentence that are relevant for predicting the next word or character of the sequence. In the attention mechanism, the context vector $c_i$  is calculated as the weighted combination of all the encoder hidden states (see Eq.\ref{attention}) and $\alpha$ refers to how much importance should be given to respective input states.

\begin{equation}
    \begin{split}
        c_i = \sum_{j=1}^{T_x} \alpha_{ij}h_j \\ 
        \alpha_{ij} = \dfrac{exp(e_{ij})}{\sum_{k=1}^{T_x} exp(e_{ik})} \\
        e_{ij} = a(s_{i-1},h_j)
    \end{split}
    \label{attention}
\end{equation}

A majority of the work done for the task of language generation was done using the encoder-decoder architecture. Zhang and Lapata \citeyearpar{zhang2014chinese} proposed a model for Chinese poetry generation with the help of RNN. In their work, they combined the process of content determination and realization was jointly into one joint process. 


Another example was the NLG system developed by Wen \emph{et al.}  \citeyearpar{wen2015semantically}, who modified the architecture of the LSTM to constrain it semantically and be able to predict the next utterance in a dialogue context. The architecture of the modified LSTM cell was used for surface realization and the dialogue act cell which acts similar to the memory cell was used for the sentence planning phase. Along similar lines was the work by Goyal \emph{et al.}  \citeyearpar{goyal2016natural}, who presented a character-level RNN for dialogue generation and addressed the issue of delexicalization and sparsity. Mei \emph{et al.}  \citeyearpar{mei2015talk} used the encoder-decoder aligner architecture 
to perform the task of content selection and realization on a set of weather database event records as a joint task. The aligner is based on the attention mechanism \citep{xu2015show,bahdanau2014neural}. The encoder-decoder architecture was also used to generate emotional text as demonstrated by the work done by Asghar \emph{et al.}, \citeyearpar{DBLP:journals/corr/abs-1709-03968}, Zhou \emph{et al.}, \citeyearpar{zhou2018emotional} and Ke \emph{et al.}, \citeyearpar{ke2018generating}. 
 
\subsection{Memory Networks}
Memory networks, a type of learning model, were introduced by Weston \emph{et al.} (\citeyear{memorynetworks2014weston}) to overcome to short memory encoded in the hidden states. These networks were used for a variety of question-answering tasks where the answer is generated from a set of facts fed into the model. The answer generated by the model can be a one-word answer or paragraph of text. The memory networks introduced by Weston \emph{et al.} \citeyearpar{memorynetworks2014weston} had four major components: input feature map, generalisation, output feature map and a response. Kumar \emph{et al.}  \citeyearpar{kumar2016ask} introduced a different type of memory network, based on episodic memory and were able to solve a wider range of question answering tasks and also on questions related to part of speech and sentiment analysis. The work done by Kumar \emph{et al.} (\citeyear{kumar2016ask}) was extended for visual question answering by Xiong \emph{et al.} (\citeyear{xiong2016dynamic}). Other works on visual question answering included the work done on using hierarchical attention on question-image pairs \citep{lu2016hierarchical}), using relational networks for generating answers for visual question answering \citep{santoro2017simple} and using facts for visual question answering \citep{wang2018fvqa}. 

\subsection{Transformer Models}
\begin{figure}[h]
    \centering
    \includegraphics[height=10cm]{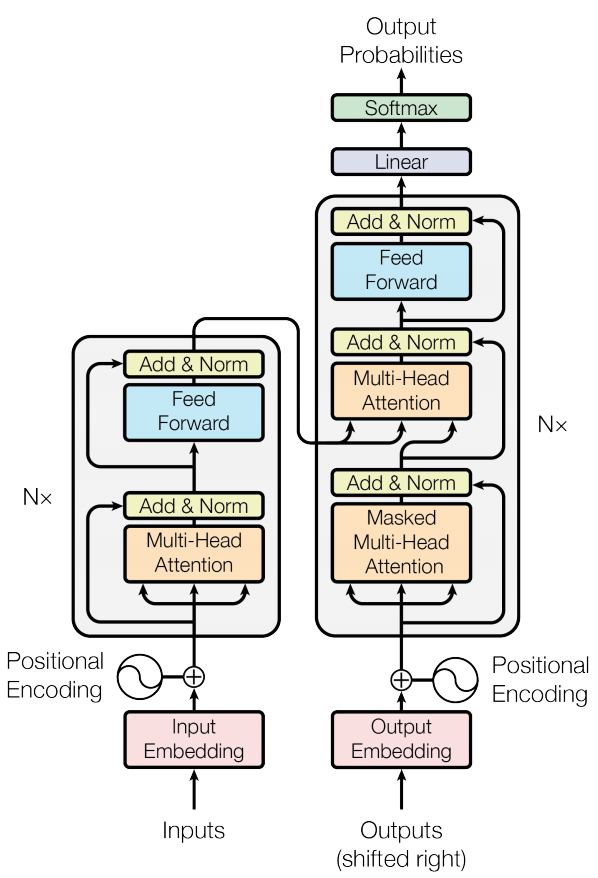}
    \caption{Transformer architecture as represented in Vaswani \emph{et al.} (Figure credit: Vaswani \emph{et al.} \citeyearpar{vaswani2017attention}}
    \label{fig:transformer}
\end{figure}

The Transformer models (Figure \ref{fig:transformer}) introduced by Vaswani \emph{et al.}  \citeyearpar{vaswani2017attention} have helped achieve improvements over a wide range of NLP tasks. Transformer models are based on attention mechanisms, that draw global dependencies between the input and output. The transformer is a made of the encoder-decoder architecture but each encoder is a stack of six encoders with each encoder containing a self-attention and point-wise fully connected feed forward neural networks. The decoder is also a stack of six decoders with each decocder containing the same components as the encoder, but with an additional attention layer that helps the decoder focus on relevant parts of the input sentence. Work using transformer models is still in its infancy. Radford \emph{et al.}, \citeyearpar{radford2018improving} and Devlin \emph{et al.} \citeyearpar{devlin2018bert} showed impressive results on several NLP tasks. 
Their work improved the existing state-of-the-art across a wide range of tasks such as language modeling, children`s book test, reading comprehension, machine translation, question answering, modeling long range dependencies (LAMBADA), Winograd Schema challenge and summarization.

\section{Open Domain Dialogue Systems using Deep Learning}
\label{DS}

Dialogue systems or conversational agents (CA) are designed with the intention of generating meaningful and coherent responses that are easy to respond to and informative when the system is engaged in a conversation with humans. A good dialogue model incorporated in conversational agents should be able to generate dialogues with high similarity to how humans converse \citep{li2017adversarial}. Conversational agents are of great importance to a large variety of applications and can be grouped under two major categories, namely, (1) Closed Domain goal-oriented systems that help users achieve a particular goal, (2) Open Domain conversational agents engaging in a conversation with a human \textendash also referred to as chit-chat models. Work on building end-to-end systems using neural networks \citep{vinyals2015neural,shang2015neural} has been increasingly published in recent years, and is the primary focus of this section.


With the fast paced advancement of research in this area, we find there is a lack of a comprehensive survey particularly in the area of the \textbf{open domain dialogue systems}. To address this gap in research, we summarize research done in this field by analyzing all the papers published in top conferences from 2015. We focus on the key aspects of these papers to summarize current trends: \textbf{corpora used, architecture implemented, optimization strategy used, evaluation metrics to evaluate efficacy}. 

These are summarized in the columns in Table \ref{Survey} and we observe the following trends: 

\textbf{Corpora} refers to the language data that has been used in the paper. The most commonly used corpora are \textit{Open Subtitles, Twitter Conversation Dialogues, Movie Triples, Cornell Movie Dialogues}. More recently, new datasets such as \textit{PERSONA chat dataset, Reddit dataset} have been made available to the community. 

\textbf{Architecture} gives an overview of the type of architecture used in the paper. Most of the research done in this field, have used variation of \textit{seq2seq} models with attention mechanism. More recently, with the creation of the transformer models, researchers have started using this architecture for the open domain dialogue systems but the work is still in its infancy.  

\textbf{Evaluation} is one of the most important aspect of open-domain dialogue systems. This is still an open research problem as there exist no appropriate or standardized metrics for evaluating performance. Researchers have primarily relied on adapting automated metrics such as BLEU \citep{papineni2002bleu}, METEOR  \citep{banerjee2005meteor} and embedding-based metrics \cite{shen2017conditional,lowe2017towards} for validating the performance. However, research has shown that these metrics show little to no correlation with evaluation from humans \citep{novikova2017we,lowe2017towards}. We find that human evaluation is another primary evaluation metric that exists in this field and researchers use different criteria such as \textit{Semantic Relevance, Appropriateness, Interestingness, Fluency, Grammar}. These are listed in Table 1. When there is no criteria presented in the paper, we simply list Human Evaluation. This refers to the cases where the researchers asked humans to simply judge which response is better between the generated and the ground-truth response.

\begin{footnotesize}
\begin{longtable}{@{}p{0.21\textwidth}p{0.22\textwidth}p{0.15\textwidth}p{0.13\textwidth}p{0.22\textwidth}@{}}
\caption{Summary of deep learning-based open-domain dialogue systems (from 2015 to present) providing an overview of the corpus used, architecture and optimization strategy implemented and evaluation metrics used in the paper.}
\footnotesize
\label{Survey}\\
\toprule
Authors        & Corpora                                                                                       & \begin{tabular}[c]{@{}l@{}}Architecture\end{tabular}                                                      & \begin{tabular}[c]{@{}l@{}}Optimization\end{tabular}                                            & \begin{tabular}[c]{@{}l@{}}Evaluation Metrics\end{tabular}                                             \\ \midrule
 \begin{tabular}[l]{@{}l@{}}\cite{vinyals2015neural} \end{tabular}  &     \begin{minipage}[t]{\linewidth}
    \begin{itemize}[wide, labelwidth=!, labelindent=0pt]
     \item \begin{tabular}[c]{@{}l@{}}Open Subtitles\end{tabular}
     \item \begin{tabular}[c]{@{}l@{}}IT Help Desk\end{tabular}
    \end{itemize}
    \end{minipage}  & \begin{minipage}[t]{\linewidth}
    \begin{itemize}[wide, labelwidth=!, labelindent=0pt]
     \item \begin{tabular}[c]{@{}l@{}}Seq2Seq\end{tabular}
    \end{itemize}
    \end{minipage} & \begin{minipage}[t]{\linewidth}
    \begin{itemize}[wide, labelwidth=!, labelindent=0pt]
     \item \begin{tabular}[c]{@{}l@{}}Cross \\Entropy\end{tabular}
    \end{itemize}
    \end{minipage} & \begin{minipage}[t]{\linewidth}
    \begin{itemize}[wide, labelwidth=!, labelindent=0pt]
     \item \begin{tabular}[c]{@{}l@{}}Human Evaluation\end{tabular}
    \end{itemize}
    \end{minipage}  \\ \midrule
 
 \begin{tabular}[l]{@{}l@{}}\cite{sordoni2015neural} \end{tabular} & \begin{minipage}[t]{\linewidth}
    \begin{itemize}[wide, labelwidth=!, labelindent=0pt]
     \item \begin{tabular}[c]{@{}l@{}}Twitter Conversation \\Dialogue\end{tabular}
    \end{itemize}
    \end{minipage} & \begin{minipage}[t]{\linewidth}
    \begin{itemize}[wide, labelwidth=!, labelindent=0pt]
     \item \begin{tabular}[c]{@{}l@{}}Language \\ Model\end{tabular}
    \end{itemize}
    \end{minipage}& \begin{minipage}[t]{\linewidth}
    \begin{itemize}[wide, labelwidth=!, labelindent=0pt]
     \item \begin{tabular}[c]{@{}l@{}}Adgrad\end{tabular}
    \end{itemize}
    \end{minipage}                                                    & \begin{minipage}[t]{\linewidth}
    \begin{itemize}[wide, labelwidth=!, labelindent=0pt]
     \item \begin{tabular}[c]{@{}l@{}}BLEU\end{tabular}
     \item \begin{tabular}[c]{@{}l@{}}METEOR\end{tabular}
     \item \begin{tabular}[c]{@{}l@{}}Human Evaluation\end{tabular}
    \end{itemize}
    \end{minipage}   \\ \midrule
 
 \begin{tabular}[l]{@{}l@{}} \cite{li2015diversity} \end{tabular}      & \begin{minipage}[t]{\linewidth}
    \begin{itemize}[wide, labelwidth=!, labelindent=0pt]
     \item \begin{tabular}[c]{@{}l@{}}Twitter Conversation \\Dialogues\end{tabular}
     \item \begin{tabular}[c]{@{}l@{}}Open Subtitles\end{tabular}
    \end{itemize}
    \end{minipage} & \begin{minipage}[t]{\linewidth}
    \begin{itemize}[wide, labelwidth=!, labelindent=0pt]
     \item \begin{tabular}[c]{@{}l@{}}Seq2Seq\end{tabular}
    \end{itemize}
    \end{minipage}                                                         & \begin{minipage}[t]{\linewidth}
    \begin{itemize}[wide, labelwidth=!, labelindent=0pt]
     \item \begin{tabular}[c]{@{}l@{}}SGD\end{tabular}
    \end{itemize}
    \end{minipage}                                                    & \begin{minipage}[t]{\linewidth}
    \begin{itemize}[wide, labelwidth=!, labelindent=0pt]
     \item \begin{tabular}[c]{@{}l@{}}BLEU\end{tabular}
     \item \begin{tabular}[c]{@{}l@{}}Distinct-1\end{tabular}
     \item \begin{tabular}[c]{@{}l@{}}Distinct-2\end{tabular}
     \item \begin{tabular}[c]{@{}l@{}}Human Evaluation\end{tabular}
    \end{itemize}
    \end{minipage} \\ \midrule

 \begin{tabular}[l]{@{}l@{}} \cite{shang2015neural} \end{tabular}  & \begin{minipage}[t]{\linewidth}
    \begin{itemize}[wide, labelwidth=!, labelindent=0pt]
     \item \begin{tabular}[c]{@{}l@{}}Weibo Conversation\end{tabular}
    \end{itemize}
    \end{minipage}& \begin{minipage}[t]{\linewidth}
    \begin{itemize}[wide, labelwidth=!, labelindent=0pt]
     \item \begin{tabular}[c]{@{}l@{}}Seq2Sseq +\\ Attention\end{tabular}
    \end{itemize}
    \end{minipage}& N/A                                                     & \begin{minipage}[t]{\linewidth}
    \begin{itemize}[wide, labelwidth=!, labelindent=0pt]
     \item \begin{tabular}[c]{@{}l@{}}Human Evaluation\end{tabular}
     \begin{itemize}[wide, labelwidth=!, labelindent=0pt]
         \item \begin{tabular}[c]{@{}l@{}}Grammar \& Fluency\end{tabular}
         \item \begin{tabular}[c]{@{}l@{}}Logic Consistency\end{tabular}
         \item \begin{tabular}[c]{@{}l@{}}Semantic Relevance\end{tabular}
         \item \begin{tabular}[c]{@{}l@{}}Scenario Dependance\end{tabular}
         \item \begin{tabular}[c]{@{}l@{}}Generality\end{tabular}
     \end{itemize}
    \end{itemize}
    \end{minipage}\\ \midrule

  \begin{tabular}[l]{@{}l@{}}\cite{yao2015attention} \end{tabular} & \begin{minipage}[t]{\linewidth}
    \begin{itemize}[wide, labelwidth=!, labelindent=0pt]
     \item \begin{tabular}[c]{@{}l@{}}Helpdesk chat service\end{tabular}
    \end{itemize}
    \end{minipage} & \begin{minipage}[t]{\linewidth}
    \begin{itemize}[wide, labelwidth=!, labelindent=0pt]
     \item \begin{tabular}[c]{@{}l@{}}Seq2Seq + \\ Attention + \\ Intention \\Network \end{tabular}
    \end{itemize}
    \end{minipage}& N/A &  \begin{minipage}[t]{\linewidth}
    \begin{itemize}[wide, labelwidth=!, labelindent=0pt]
     \item \begin{tabular}[c]{@{}l@{}}Perplexity\end{tabular}
    \end{itemize}
    \end{minipage}\\ \midrule
  
   \begin{tabular}[l]{@{}l@{}}\cite{serban2016building} \end{tabular} &\begin{minipage}[t]{\linewidth}
    \begin{itemize}[wide, labelwidth=!, labelindent=0pt]
     \item \begin{tabular}[c]{@{}l@{}}Movie Triples\end{tabular}
    \end{itemize}
    \end{minipage} & \begin{minipage}[t]{\linewidth}
    \begin{itemize}[wide, labelwidth=!, labelindent=0pt]
     \item \begin{tabular}[c]{@{}l@{}}Hierarchical\\ Encoder \\ Decoder\end{tabular}
    \end{itemize}
    \end{minipage}& \begin{minipage}[t]{\linewidth}
    \begin{itemize}[wide, labelwidth=!, labelindent=0pt]
     \item \begin{tabular}[c]{@{}l@{}}Adam\end{tabular}
    \end{itemize}
    \end{minipage} &  \begin{minipage}[t]{\linewidth}
    \begin{itemize}[wide, labelwidth=!, labelindent=0pt]
     \item \begin{tabular}[c]{@{}l@{}}Perplexity\end{tabular}
    \end{itemize}
    \end{minipage}\\ \midrule
   
    \begin{tabular}[l]{@{}l@{}}\cite{li2016persona} \end{tabular} & \begin{minipage}[t]{\linewidth}
    \begin{itemize}[wide, labelwidth=!, labelindent=0pt]
     \item \begin{tabular}[c]{@{}l@{}}Twitter Persona \\dataset\end{tabular}
     \item \begin{tabular}[c]{@{}l@{}}Twitter Conversation \\ Dialogues\end{tabular}
     \item \begin{tabular}[c]{@{}l@{}}TV Series Transcripts\end{tabular}
    \end{itemize}
    \end{minipage}  & \begin{minipage}[t]{\linewidth}
    \begin{itemize}[wide, labelwidth=!, labelindent=0pt]
     \item \begin{tabular}[c]{@{}l@{}}Seq2Seq + \\ Persona \\Embeddings \end{tabular}
    \end{itemize}
    \end{minipage}& \begin{minipage}[t]{\linewidth}
    \begin{itemize}[wide, labelwidth=!, labelindent=0pt]
     \item \begin{tabular}[c]{@{}l@{}}MMI \end{tabular}
    \end{itemize}
    \end{minipage} &  \begin{minipage}[t]{\linewidth}
    \begin{itemize}[wide, labelwidth=!, labelindent=0pt]
     \item \begin{tabular}[c]{@{}l@{}}Perplexity\end{tabular}
     \item \begin{tabular}[c]{@{}l@{}}BLEU\end{tabular}
     \item \begin{tabular}[c]{@{}l@{}}Human Evaluation\end{tabular}
    \end{itemize}
    \end{minipage} \\ \midrule
    
     \begin{tabular}[l]{@{}l@{}}\cite{luan2016lstm} \end{tabular} & \begin{minipage}[t]{\linewidth}
    \begin{itemize}[wide, labelwidth=!, labelindent=0pt]
     \item \begin{tabular}[c]{@{}l@{}}Ubuntu Dialogues\end{tabular}
    \end{itemize}
    \end{minipage} & \begin{minipage}[t]{\linewidth}
    \begin{itemize}[wide, labelwidth=!, labelindent=0pt]
     \item \begin{tabular}[c]{@{}l@{}}Language \\Model +  LDA \end{tabular}
    \end{itemize}
    \end{minipage}&  \begin{minipage}[t]{\linewidth}
    \begin{itemize}[wide, labelwidth=!, labelindent=0pt]
     \item \begin{tabular}[c]{@{}l@{}}SGD \end{tabular}
    \end{itemize}
    \end{minipage}  &  \begin{minipage}[t]{\linewidth}
    \begin{itemize}[wide, labelwidth=!, labelindent=0pt]
     \item \begin{tabular}[c]{@{}l@{}}Perplexity\end{tabular}
     \item \begin{tabular}[c]{@{}l@{}}Response Ranking\end{tabular}
    \end{itemize}
    \end{minipage} \\ \midrule

 \begin{tabular}[l]{@{}l@{}}\cite{li2016deep} \end{tabular} &  \begin{minipage}[t]{\linewidth}
    \begin{itemize}[wide, labelwidth=!, labelindent=0pt]
     \item \begin{tabular}[c]{@{}l@{}}Open Subtitles\end{tabular}
    \end{itemize}
    \end{minipage} & \begin{minipage}[t]{\linewidth}
    \begin{itemize}[wide, labelwidth=!, labelindent=0pt]
     \item \begin{tabular}[c]{@{}l@{}}Seq2Seq + RL \end{tabular}
    \end{itemize}
    \end{minipage}&  \begin{minipage}[t]{\linewidth}
    \begin{itemize}[wide, labelwidth=!, labelindent=0pt]
     \item \begin{tabular}[c]{@{}l@{}}MMI +\\ Policy \\Gradient \end{tabular}
    \end{itemize}
    \end{minipage}&  \begin{minipage}[t]{\linewidth}
    \begin{itemize}[wide, labelwidth=!, labelindent=0pt]
     \item \begin{tabular}[c]{@{}l@{}}BLEU\end{tabular}
     \item \begin{tabular}[c]{@{}l@{}}Dialogue Length\end{tabular}
     \item \begin{tabular}[c]{@{}l@{}}Diversity\end{tabular}
     \item \begin{tabular}[c]{@{}l@{}}Human Evaluation\end{tabular}
    \end{itemize}
    \end{minipage} \\ \midrule

 \begin{tabular}[l]{@{}l@{}}\cite{duvsek2016context} \end{tabular} &  \begin{minipage}[t]{\linewidth}
    \begin{itemize}[wide, labelwidth=!, labelindent=0pt]
     \item \begin{tabular}[c]{@{}l@{}}Public Transport \\Information\end{tabular}
    \end{itemize}
    \end{minipage} &  \begin{minipage}[t]{\linewidth}
    \begin{itemize}[wide, labelwidth=!, labelindent=0pt]
     \item \begin{tabular}[c]{@{}l@{}}Seq2Seq + \\ Attention + \\ Context \\ Encoder \end{tabular}
    \end{itemize}
    \end{minipage}&  \begin{minipage}[t]{\linewidth}
    \begin{itemize}[wide, labelwidth=!, labelindent=0pt]
     \item \begin{tabular}[c]{@{}l@{}}Cross\\ Entropy \end{tabular}
    \end{itemize}
    \end{minipage} &  \begin{minipage}[t]{\linewidth}
    \begin{itemize}[wide, labelwidth=!, labelindent=0pt]
     \item \begin{tabular}[c]{@{}l@{}}BLEU\end{tabular}
     \item \begin{tabular}[c]{@{}l@{}}NIST\end{tabular}
     \item \begin{tabular}[c]{@{}l@{}}Human Evaluation\end{tabular}
    \end{itemize}
    \end{minipage} \\ \midrule
 
  \begin{tabular}[l]{@{}l@{}}\cite{mou2016sequence} \end{tabular}&  \begin{minipage}[t]{\linewidth}
    \begin{itemize}[wide, labelwidth=!, labelindent=0pt]
     \item \begin{tabular}[c]{@{}l@{}}Baidu Teiba Forum\end{tabular}
    \end{itemize}
    \end{minipage} &  \begin{minipage}[t]{\linewidth}
    \begin{itemize}[wide, labelwidth=!, labelindent=0pt]
     \item \begin{tabular}[c]{@{}l@{}}Seq2Seq \end{tabular}
    \end{itemize}
    \end{minipage} & \begin{minipage}[t]{\linewidth}
    \begin{itemize}[wide, labelwidth=!, labelindent=0pt]
     \item \begin{tabular}[c]{@{}l@{}}SGD\end{tabular}
    \end{itemize}
    \end{minipage} &  \begin{minipage}[t]{\linewidth}
    \begin{itemize}[wide, labelwidth=!, labelindent=0pt]
     \item \begin{tabular}[c]{@{}l@{}}Human Evaluation\end{tabular}
     \item \begin{tabular}[c]{@{}l@{}}Length\end{tabular}
     \item \begin{tabular}[c]{@{}l@{}}Entropy\end{tabular}
    \end{itemize}
    \end{minipage} \\ \midrule
  
    \begin{tabular}[l]{@{}l@{}}\cite{asghar2016deep} \end{tabular} &  \begin{minipage}[t]{\linewidth}
    \begin{itemize}[wide, labelwidth=!, labelindent=0pt]
     \item \begin{tabular}[c]{@{}l@{}}Cornell Movie \\ Dialogues\end{tabular}
    \end{itemize}
    \end{minipage}  &  \begin{minipage}[t]{\linewidth}
    \begin{itemize}[wide, labelwidth=!, labelindent=0pt]
     \item \begin{tabular}[c]{@{}l@{}}Seq2Seq + \\ Online \\ Active \\ Learning \end{tabular}
    \end{itemize}
    \end{minipage} &  \begin{minipage}[t]{\linewidth}
    \begin{itemize}[wide, labelwidth=!, labelindent=0pt]
     \item \begin{tabular}[c]{@{}l@{}}Cross \\ Entropy \end{tabular}
    \end{itemize}
    \end{minipage}  &  \begin{minipage}[t]{\linewidth}
    \begin{itemize}[wide, labelwidth=!, labelindent=0pt]
     \item \begin{tabular}[c]{@{}l@{}}Human Evaluation\end{tabular}
     \item \begin{tabular}[c]{@{}l@{}}Syntactic Coherence\end{tabular}
     \item \begin{tabular}[c]{@{}l@{}}Relevance\end{tabular}
     \item \begin{tabular}[c]{@{}l@{}}Interestingness\end{tabular}
    \end{itemize}
    \end{minipage} \\ \midrule
  
   \begin{tabular}[l]{@{}l@{}}\cite{serban2017hierarchical} \end{tabular} &  \begin{minipage}[t]{\linewidth}
    \begin{itemize}[wide, labelwidth=!, labelindent=0pt]
     \item \begin{tabular}[c]{@{}l@{}}Twitter Conversation \\Dialogues\end{tabular}
     \item \begin{tabular}[c]{@{}l@{}}Ubuntu Dialogues\end{tabular}
    \end{itemize}
    \end{minipage}& \begin{minipage}[t]{\linewidth}
    \begin{itemize}[wide, labelwidth=!, labelindent=0pt]
     \item \begin{tabular}[c]{@{}l@{}}Latent \\Variable \\ Hierarchical\\ Encoder \\ Decoder\end{tabular}
    \end{itemize}
    \end{minipage}&  \begin{minipage}[t]{\linewidth}
    \begin{itemize}[wide, labelwidth=!, labelindent=0pt]
     \item \begin{tabular}[c]{@{}l@{}}Adam\end{tabular}
    \end{itemize}
    \end{minipage} &  \begin{minipage}[t]{\linewidth}
    \begin{itemize}[wide, labelwidth=!, labelindent=0pt]
     \item \begin{tabular}[c]{@{}l@{}}Human Evaluation\end{tabular}
     \item \begin{tabular}[c]{@{}l@{}}Length\end{tabular}
     \item \begin{tabular}[c]{@{}l@{}}Entropy\end{tabular}
    \end{itemize}
    \end{minipage}  \\ \midrule

 \begin{tabular}[l]{@{}l@{}}\cite{mei2017coherent} \end{tabular} &  \begin{minipage}[t]{\linewidth}
    \begin{itemize}[wide, labelwidth=!, labelindent=0pt]
     \item \begin{tabular}[c]{@{}l@{}}Movie Triples\end{tabular}
     \item \begin{tabular}[c]{@{}l@{}}Ubuntu Dialogues\end{tabular}
    \end{itemize}
    \end{minipage}&  \begin{minipage}[t]{\linewidth}
    \begin{itemize}[wide, labelwidth=!, labelindent=0pt]
     \item \begin{tabular}[c]{@{}l@{}}Language \\ Models \\+ Attention \\ + LDA \\ Reranking\end{tabular}
    \end{itemize}
    \end{minipage} & \begin{minipage}[t]{\linewidth}
    \begin{itemize}[wide, labelwidth=!, labelindent=0pt]
     \item \begin{tabular}[c]{@{}l@{}}Adam\end{tabular}
    \end{itemize}
    \end{minipage}&  \begin{minipage}[t]{\linewidth}
    \begin{itemize}[wide, labelwidth=!, labelindent=0pt]
    \item \begin{tabular}[c]{@{}l@{}}Perplexity\end{tabular}
    \item \begin{tabular}[c]{@{}l@{}}Word Error Rate\end{tabular}
    \item \begin{tabular}[c]{@{}l@{}}Recall\end{tabular}
    \item \begin{tabular}[c]{@{}l@{}}Distinct-1\end{tabular}
     \item \begin{tabular}[c]{@{}l@{}}Human Evaluation\end{tabular}
    \end{itemize}
    \end{minipage}\\ \midrule

 \begin{tabular}[l]{@{}l@{}}\cite{xing2017topic} \end{tabular} &  \begin{minipage}[t]{\linewidth}
    \begin{itemize}[wide, labelwidth=!, labelindent=0pt]
     \item \begin{tabular}[c]{@{}l@{}}Baidu Teiba Forum\end{tabular}
    \end{itemize}
    \end{minipage}&  \begin{minipage}[t]{\linewidth}
    \begin{itemize}[wide, labelwidth=!, labelindent=0pt]
     \item \begin{tabular}[c]{@{}l@{}}Seq2Seq \\ + LDA \\ + Joint \\ Attention\end{tabular}
    \end{itemize}
    \end{minipage}& \begin{minipage}[t]{\linewidth}
    \begin{itemize}[wide, labelwidth=!, labelindent=0pt]
     \item \begin{tabular}[c]{@{}l@{}}Adadelta\end{tabular}
    \end{itemize}
    \end{minipage} &  \begin{minipage}[t]{\linewidth}
    \begin{itemize}[wide, labelwidth=!, labelindent=0pt]
    \item \begin{tabular}[c]{@{}l@{}}Perplexity\end{tabular}
    \item \begin{tabular}[c]{@{}l@{}}Distinct-1\end{tabular}
    \item \begin{tabular}[c]{@{}l@{}}Distinct-2\end{tabular}
     \item \begin{tabular}[c]{@{}l@{}}Human Evaluation\end{tabular}
    \end{itemize}
    \end{minipage}\\ \midrule
  
  \begin{tabular}[l]{@{}l@{}} \cite{cao2017latent} \end{tabular} &   \begin{minipage}[t]{\linewidth}
    \begin{itemize}[wide, labelwidth=!, labelindent=0pt]
     \item \begin{tabular}[c]{@{}l@{}}Open Subtitles\end{tabular}
    \end{itemize}
    \end{minipage} &  \begin{minipage}[t]{\linewidth}
    \begin{itemize}[wide, labelwidth=!, labelindent=0pt]
     \item \begin{tabular}[c]{@{}l@{}}Variational \\Autoencoder\end{tabular}
    \end{itemize}
    \end{minipage} &  \begin{minipage}[t]{\linewidth}
    \begin{itemize}[wide, labelwidth=!, labelindent=0pt]
     \item \begin{tabular}[c]{@{}l@{}}MMI\end{tabular}
    \end{itemize}
    \end{minipage}&   \begin{minipage}[t]{\linewidth}
    \begin{itemize}[wide, labelwidth=!, labelindent=0pt]
     \item \begin{tabular}[c]{@{}l@{}}Human Evaluation\end{tabular}
    \end{itemize}
    \end{minipage}\\ \midrule

 \begin{tabular}[l]{@{}l@{}}\cite{lewis2017deal} \end{tabular} &   \begin{minipage}[t]{\linewidth}
    \begin{itemize}[wide, labelwidth=!, labelindent=0pt]
     \item \begin{tabular}[c]{@{}l@{}}Negotiation dataset\end{tabular}
    \end{itemize}
    \end{minipage}&  \begin{minipage}[t]{\linewidth}
    \begin{itemize}[wide, labelwidth=!, labelindent=0pt]
     \item \begin{tabular}[c]{@{}l@{}}Seq2Seq + \\ self play +\\ RL\end{tabular}
    \end{itemize}
    \end{minipage}& \begin{minipage}[t]{\linewidth}
    \begin{itemize}[wide, labelwidth=!, labelindent=0pt]
     \item \begin{tabular}[c]{@{}l@{}}SGD\end{tabular}
    \end{itemize}
    \end{minipage}   &   \begin{minipage}[t]{\linewidth}
    \begin{itemize}[wide, labelwidth=!, labelindent=0pt]
     \item \begin{tabular}[c]{@{}l@{}}Human Evaluation\end{tabular}
     \begin{itemize}[wide, labelwidth=!, labelindent=0pt]
         \item \begin{tabular}[c]{@{}l@{}}Score\end{tabular}
         \item \begin{tabular}[c]{@{}l@{}}Agreement\end{tabular}
         \item \begin{tabular}[c]{@{}l@{}}Pareto Optimality\end{tabular}
     \end{itemize}
    \item \begin{tabular}[c]{@{}l@{}}Perplexity\end{tabular}
    \end{itemize}
    \end{minipage} \\ \midrule
 
  \begin{tabular}[l]{@{}l@{}}\cite{li2017adversarial} \end{tabular} &  \begin{minipage}[t]{\linewidth}
    \begin{itemize}[wide, labelwidth=!, labelindent=0pt]
     \item \begin{tabular}[c]{@{}l@{}}Open Subtitles\end{tabular}
    \end{itemize}
    \end{minipage} & \begin{minipage}[t]{\linewidth}
    \begin{itemize}[wide, labelwidth=!, labelindent=0pt]
     \item \begin{tabular}[c]{@{}l@{}}GAN\end{tabular}
    \end{itemize}
    \end{minipage}& N/A  &  \begin{minipage}[t]{\linewidth}
    \begin{itemize}[wide, labelwidth=!, labelindent=0pt]
     \item \begin{tabular}[c]{@{}l@{}}Human Evaluation\end{tabular}
     \item \begin{tabular}[c]{@{}l@{}}Adversarial Evaluation\end{tabular}
    \end{itemize}
    \end{minipage}   \\ \midrule

 \begin{tabular}[l]{@{}l@{}} \cite{qian2017assigning} \end{tabular} &  \begin{minipage}[t]{\linewidth}
    \begin{itemize}[wide, labelwidth=!, labelindent=0pt]
     \item \begin{tabular}[c]{@{}l@{}}Weibo Dataset\end{tabular}
     \item \begin{tabular}[c]{@{}l@{}}Profile Binary Subset\end{tabular}
     \item \begin{tabular}[c]{@{}l@{}}Profile Related Subset\end{tabular}
     \item \begin{tabular}[c]{@{}l@{}}Manual Dataset\end{tabular}
    \end{itemize}
    \end{minipage}  & \begin{minipage}[t]{\linewidth}
    \begin{itemize}[wide, labelwidth=!, labelindent=0pt]
     \item \begin{tabular}[c]{@{}l@{}}Encoder\\ Decoder + \\ Profile \\ Detector\end{tabular}
    \end{itemize}
    \end{minipage} & \begin{minipage}[t]{\linewidth}
    \begin{itemize}[wide, labelwidth=!, labelindent=0pt]
     \item \begin{tabular}[c]{@{}l@{}}SGD\end{tabular}
    \end{itemize}
    \end{minipage}&  \begin{minipage}[t]{\linewidth}
    \begin{itemize}[wide, labelwidth=!, labelindent=0pt]
     \item \begin{tabular}[c]{@{}l@{}}Human Evaluation\end{tabular}
     \begin{itemize}[wide, labelwidth=!, labelindent=0pt]
         \item \begin{tabular}[c]{@{}l@{}}Naturalness\end{tabular}
         \item \begin{tabular}[c]{@{}l@{}}Logic\end{tabular}
         \item \begin{tabular}[c]{@{}l@{}}Semantic\\Relevance\end{tabular}
         \item \begin{tabular}[c]{@{}l@{}}Correctness\end{tabular}
         \item \begin{tabular}[c]{@{}l@{}}Consistency\end{tabular}
         \item \begin{tabular}[c]{@{}l@{}}Variety\end{tabular}
     \end{itemize}
     \item \begin{tabular}[c]{@{}l@{}}Profile Detection\end{tabular}
     \item \begin{tabular}[c]{@{}l@{}}Position Detection\end{tabular}
    \end{itemize}
    \end{minipage}
 \\ \midrule

 \begin{tabular}[l]{@{}l@{}}\cite{qiu2017alime} \end{tabular} &  \begin{minipage}[t]{\linewidth}
    \begin{itemize}[wide, labelwidth=!, labelindent=0pt]
     \item \begin{tabular}[c]{@{}l@{}}Chatlog Online \\ Customer Service\end{tabular}
    \end{itemize}
    \end{minipage}&  \begin{minipage}[t]{\linewidth}
    \begin{itemize}[wide, labelwidth=!, labelindent=0pt]
     \item \begin{tabular}[c]{@{}l@{}}Attentive \\ Seq2Seq \\ + IR \\+ Rerank\end{tabular}
    \end{itemize}
    \end{minipage}& N/A  &  \begin{minipage}[t]{\linewidth}
    \begin{itemize}[wide, labelwidth=!, labelindent=0pt]
     \item \begin{tabular}[c]{@{}l@{}}Precision\end{tabular}
     \item \begin{tabular}[c]{@{}l@{}}Recall\end{tabular}
     \item \begin{tabular}[c]{@{}l@{}}F1 score\end{tabular}
     \item \begin{tabular}[c]{@{}l@{}}Human Evaluation\end{tabular}
    \end{itemize}
    \end{minipage}\\ \midrule

 \begin{tabular}[l]{@{}l@{}} \cite{serban2017multiresolution} \end{tabular} &  \begin{minipage}[t]{\linewidth}
    \begin{itemize}[wide, labelwidth=!, labelindent=0pt]
     \item \begin{tabular}[c]{@{}l@{}}Ubuntu Dialogues \end{tabular}
     \item \begin{tabular}[c]{@{}l@{}}Twitter Conversation \\ Dialogues \end{tabular}
    \end{itemize}
    \end{minipage}&  \begin{minipage}[t]{\linewidth}
    \begin{itemize}[wide, labelwidth=!, labelindent=0pt]
     \item \begin{tabular}[c]{@{}l@{}}MrRNN\end{tabular}
    \end{itemize}
    \end{minipage}&  \begin{minipage}[t]{\linewidth}
    \begin{itemize}[wide, labelwidth=!, labelindent=0pt]
     \item \begin{tabular}[c]{@{}l@{}}Adam\end{tabular}
    \end{itemize}
    \end{minipage} & \begin{minipage}[t]{\linewidth}
    \begin{itemize}[wide, labelwidth=!, labelindent=0pt]
     \item \begin{tabular}[c]{@{}l@{}}Human Evaluation \end{tabular}
    \end{itemize}
    \end{minipage}\\ \midrule

 \begin{tabular}[l]{@{}l@{}}\cite{shen2017conditional} \end{tabular} &  \begin{minipage}[t]{\linewidth}
    \begin{itemize}[wide, labelwidth=!, labelindent=0pt]
     \item \begin{tabular}[c]{@{}l@{}}Ubuntu Dialogues \end{tabular}
    \end{itemize}
    \end{minipage}  &  \begin{minipage}[t]{\linewidth}
    \begin{itemize}[wide, labelwidth=!, labelindent=0pt]
     \item \begin{tabular}[c]{@{}l@{}}Hierarchical\\ Encoder\\ Decoder\end{tabular}
    \end{itemize}
    \end{minipage} &  \begin{minipage}[t]{\linewidth}
    \begin{itemize}[wide, labelwidth=!, labelindent=0pt]
     \item \begin{tabular}[c]{@{}l@{}}KL \\Divergence\end{tabular}
    \end{itemize}
    \end{minipage}&  \begin{minipage}[t]{\linewidth}
    \begin{itemize}[wide, labelwidth=!, labelindent=0pt]
     \item \begin{tabular}[c]{@{}l@{}}Human Evaluation\end{tabular}
     \begin{itemize}[wide, labelwidth=!, labelindent=0pt]
         \item \begin{tabular}[c]{@{}l@{}}Grammaticality\end{tabular}
         \item \begin{tabular}[c]{@{}l@{}}Coherence\end{tabular}
         \item \begin{tabular}[c]{@{}l@{}}Diversity\end{tabular}
     \end{itemize}
    \item \begin{tabular}[c]{@{}l@{}}Embedding Evaluation\end{tabular}
     \begin{itemize}[wide, labelwidth=!, labelindent=0pt]
         \item \begin{tabular}[c]{@{}l@{}}Greedy\end{tabular}
         \item \begin{tabular}[c]{@{}l@{}}Average\end{tabular}
         \item \begin{tabular}[c]{@{}l@{}}Extrema\end{tabular}
     \end{itemize}
    \end{itemize}
    \end{minipage} \\ \midrule
 
 \begin{tabular}[l]{@{}l@{}}\cite{tian2017make} \end{tabular} & \begin{minipage}[t]{\linewidth}
    \begin{itemize}[wide, labelwidth=!, labelindent=0pt]
     \item \begin{tabular}[c]{@{}l@{}}Baidu Teiba Forum\end{tabular}
    \end{itemize}
    \end{minipage} &  \begin{minipage}[t]{\linewidth}
    \begin{itemize}[wide, labelwidth=!, labelindent=0pt]
     \item \begin{tabular}[c]{@{}l@{}}Hierarchical\\ Encoder\\ Decoder\end{tabular}
    \end{itemize}
    \end{minipage} &  \begin{minipage}[t]{\linewidth}
    \begin{itemize}[wide, labelwidth=!, labelindent=0pt]
     \item \begin{tabular}[c]{@{}l@{}}AdaDelta\end{tabular}
    \end{itemize}
    \end{minipage} &  \begin{minipage}[t]{\linewidth}
    \begin{itemize}[wide, labelwidth=!, labelindent=0pt]
     \item \begin{tabular}[c]{@{}l@{}}BLEU\end{tabular}
     \item \begin{tabular}[c]{@{}l@{}}Length\end{tabular}
     \item \begin{tabular}[c]{@{}l@{}}Entropy\end{tabular}
     \item \begin{tabular}[c]{@{}l@{}}Diversity\end{tabular}
    \end{itemize}
    \end{minipage}\\ \midrule
 
  \begin{tabular}[l]{@{}l@{}}\cite{bhatia2017soc2seq} \end{tabular} &  \begin{minipage}[t]{\linewidth}
    \begin{itemize}[wide, labelwidth=!, labelindent=0pt]
     \item \begin{tabular}[c]{@{}l@{}}Yik Yak Dataset\end{tabular}
    \end{itemize}
    \end{minipage}&  \begin{minipage}[t]{\linewidth}
    \begin{itemize}[wide, labelwidth=!, labelindent=0pt]
     \item \begin{tabular}[c]{@{}l@{}}Seq2Seq \\+ Locations\end{tabular}
     \item \begin{tabular}[l]{@{}l@{}} Seq2Seq \\+ User \\model\end{tabular}
    \end{itemize}
    \end{minipage}&  \begin{tabular}[l]{@{}l@{}} N/A \end{tabular}&  \begin{minipage}[t]{\linewidth}
    \begin{itemize}[wide, labelwidth=!, labelindent=0pt]
     \item \begin{tabular}[c]{@{}l@{}}Perplexity\end{tabular}
     \item \begin{tabular}[c]{@{}l@{}}ROUGE\end{tabular}
    \end{itemize}
    \end{minipage}\\ \midrule
  
   \begin{tabular}[l]{@{}l@{}}\cite{ghosh2017affect} \end{tabular} &  \begin{minipage}[t]{\linewidth}
    \begin{itemize}[wide, labelwidth=!, labelindent=0pt]
     \item \begin{tabular}[c]{@{}l@{}}Fisher English \\ Training Speech Parts\end{tabular}
     \item \begin{tabular}[c]{@{}l@{}}Distress Assessment \\ Interview\end{tabular}
     \item \begin{tabular}[c]{@{}l@{}}SEMAINE Dataset \end{tabular}
     \item \begin{tabular}[c]{@{}l@{}}CMU-MOSI Dataset\end{tabular}
    \end{itemize}
    \end{minipage} &  \begin{minipage}[t]{\linewidth}
    \begin{itemize}[wide, labelwidth=!, labelindent=0pt]
     \item \begin{tabular}[c]{@{}l@{}}Language \\Model\end{tabular}
    \end{itemize}
    \end{minipage}& N/A &  \begin{minipage}[t]{\linewidth}
    \begin{itemize}[wide, labelwidth=!, labelindent=0pt]
     \item \begin{tabular}[c]{@{}l@{}}Perplexity\end{tabular}
     \item \begin{tabular}[c]{@{}l@{}}Human Evaluation\end{tabular}
    \end{itemize}
    \end{minipage}\\ \midrule
 
  \begin{tabular}[l]{@{}l@{}}\cite{kottur2017exploring} \end{tabular} &  \begin{minipage}[t]{\linewidth}
    \begin{itemize}[wide, labelwidth=!, labelindent=0pt]
     \item \begin{tabular}[c]{@{}l@{}}Movies-DiC Dataset\end{tabular}
     \item \begin{tabular}[c]{@{}l@{}}TV Series Transcripts\end{tabular}
     \item \begin{tabular}[c]{@{}l@{}}Open Subtitles\end{tabular}
    \end{itemize}
    \end{minipage}&  \begin{minipage}[t]{\linewidth}
    \begin{itemize}[wide, labelwidth=!, labelindent=0pt]
     \item \begin{tabular}[c]{@{}l@{}}Context-\\ aware\\ Persona \\based \\ Hierarchical \\ Encoder\\Decoder\end{tabular}
    \end{itemize}
    \end{minipage}&  \begin{minipage}[t]{\linewidth}
    \begin{itemize}[wide, labelwidth=!, labelindent=0pt]
     \item \begin{tabular}[c]{@{}l@{}}Adam\end{tabular}
    \end{itemize}
    \end{minipage}&  \begin{minipage}[t]{\linewidth}
    \begin{itemize}[wide, labelwidth=!, labelindent=0pt]
     \item \begin{tabular}[c]{@{}l@{}}Perplexity\end{tabular}
     \item \begin{tabular}[c]{@{}l@{}}Recall@1\end{tabular}
     \item \begin{tabular}[c]{@{}l@{}}Recall@5\end{tabular}
    \end{itemize}
    \end{minipage} \\ \midrule
  
  \begin{tabular}[l]{@{}l@{}}\cite{xing2018hierarchical} \end{tabular} &  \begin{minipage}[t]{\linewidth}
    \begin{itemize}[wide, labelwidth=!, labelindent=0pt]
     \item \begin{tabular}[c]{@{}l@{}}Douban Group Dataset\end{tabular}
    \end{itemize}
    \end{minipage} &  \begin{minipage}[t]{\linewidth}
    \begin{itemize}[wide, labelwidth=!, labelindent=0pt]
     \item \begin{tabular}[c]{@{}l@{}}Hierarchical \\ Recurrent \\ Attention \\ Network\end{tabular}
    \end{itemize}
    \end{minipage} &  \begin{tabular}[l]{@{}l@{}}N/A \end{tabular}  & \begin{minipage}[t]{\linewidth}
    \begin{itemize}[wide, labelwidth=!, labelindent=0pt]
     \item \begin{tabular}[c]{@{}l@{}}Perplexity\end{tabular}
     \item \begin{tabular}[c]{@{}l@{}}Human Evaluation\end{tabular}
    \end{itemize}
    \end{minipage}\\ \midrule
  
  \begin{tabular}[l]{@{}l@{}}\cite{zhou2018emotional} \end{tabular} &  \begin{minipage}[t]{\linewidth}
    \begin{itemize}[wide, labelwidth=!, labelindent=0pt]
     \item \begin{tabular}[c]{@{}l@{}}NLPCC Dataset\end{tabular}
     \item \begin{tabular}[c]{@{}l@{}}STC Dataset\end{tabular}
     \item \begin{tabular}[c]{@{}l@{}}Weibo Emotion \\Dataset\end{tabular}
    \end{itemize}
    \end{minipage} & \begin{minipage}[t]{\linewidth}
    \begin{itemize}[wide, labelwidth=!, labelindent=0pt]
     \item \begin{tabular}[c]{@{}l@{}}Encoder \\ Decoder + \\ External \\ Memory +\\ Internal \\Memory + \\ Emotion \\Embedding\end{tabular}
    \end{itemize}
    \end{minipage}  & \begin{minipage}[t]{\linewidth}
    \begin{itemize}[wide, labelwidth=!, labelindent=0pt]
     \item \begin{tabular}[c]{@{}l@{}}Cross \\ Entropy \end{tabular}
    \end{itemize}
    \end{minipage}  &  \begin{minipage}[t]{\linewidth}
    \begin{itemize}[wide, labelwidth=!, labelindent=0pt]
     \item \begin{tabular}[c]{@{}l@{}}Human Evaluation\end{tabular}
     \begin{itemize}[wide, labelwidth=!, labelindent=0pt]
         \item \begin{tabular}[c]{@{}l@{}}Content\end{tabular}
         \item \begin{tabular}[c]{@{}l@{}}Emotion\end{tabular}
     \end{itemize}
     \item \begin{tabular}[c]{@{}l@{}}Perplexity\end{tabular}
     \item \begin{tabular}[c]{@{}l@{}}Accuracy\end{tabular}
    \end{itemize}
    \end{minipage} \\ \midrule

 \begin{tabular}[l]{@{}l@{}}\cite{asghar2018affective} \end{tabular} &  \begin{minipage}[t]{\linewidth}
    \begin{itemize}[wide, labelwidth=!, labelindent=0pt]
     \item \begin{tabular}[c]{@{}l@{}}Cornell Movie\\Dialogues\end{tabular}
    \end{itemize}
    \end{minipage}&  \begin{minipage}[t]{\linewidth}
    \begin{itemize}[wide, labelwidth=!, labelindent=0pt]
     \item \begin{tabular}[l]{@{}l@{}} Seq2Seq + \\ Affective \\ Embeddings \end{tabular}
    \end{itemize}
    \end{minipage}&  \begin{minipage}[t]{\linewidth}
    \begin{itemize}[wide, labelwidth=!, labelindent=0pt]
     \item \begin{tabular}[l]{@{}l@{}} Cross \\ Entropy \end{tabular}
     \item \begin{tabular}[l]{@{}l@{}} Min \\ Affective \\ Dissonance \end{tabular}
     \item \begin{tabular}[l]{@{}l@{}} Max \\ Affective \\ Dissonance \end{tabular}
     \item \begin{tabular}[l]{@{}l@{}} Max \\ Affective \\ Content \end{tabular}
    \end{itemize}
    \end{minipage} &  \begin{minipage}[t]{\linewidth}
    \begin{itemize}[wide, labelwidth=!, labelindent=0pt]
     \item \begin{tabular}[c]{@{}l@{}}Human Evaluation\end{tabular}
     \begin{itemize}[wide, labelwidth=!, labelindent=0pt]
         \item \begin{tabular}[c]{@{}l@{}}Syntactic Coherence\end{tabular}
         \item \begin{tabular}[c]{@{}l@{}}Naturalness\end{tabular}
         \item \begin{tabular}[c]{@{}l@{}}Emotional \\ Appropriateness\end{tabular}
     \end{itemize}
    \end{itemize}
    \end{minipage}\\ \midrule

 \begin{tabular}[l]{@{}l@{}}\cite{zhang2018learning} \end{tabular} &  \begin{minipage}[t]{\linewidth}
    \begin{itemize}[wide, labelwidth=!, labelindent=0pt]
     \item \begin{tabular}[c]{@{}l@{}}STC Dataset\end{tabular}
    \end{itemize}
    \end{minipage}&  \begin{minipage}[t]{\linewidth}
    \begin{itemize}[wide, labelwidth=!, labelindent=0pt]
     \item \begin{tabular}[l]{@{}l@{}} Specificity \\Controlled \\ Seq2Seq \end{tabular}
    \end{itemize}
    \end{minipage} &  \begin{minipage}[t]{\linewidth}
    \begin{itemize}[wide, labelwidth=!, labelindent=0pt]
     \item \begin{tabular}[l]{@{}l@{}} Adam\end{tabular}
    \end{itemize}
    \end{minipage}&  \begin{minipage}[t]{\linewidth}
    \begin{itemize}[wide, labelwidth=!, labelindent=0pt]
     \item \begin{tabular}[l]{@{}l@{}} BLEU-1 \end{tabular}
     \item \begin{tabular}[l]{@{}l@{}} BLEU-2 \end{tabular}
     \item \begin{tabular}[l]{@{}l@{}} Distinct-1 \end{tabular}
     \item \begin{tabular}[l]{@{}l@{}} Distinct-2 \end{tabular}
     \item \begin{tabular}[l]{@{}l@{}} Average Embedding\end{tabular}
     \item \begin{tabular}[l]{@{}l@{}} Extrema Embedding\end{tabular}
    \end{itemize}
    \end{minipage}\\ \midrule
    
      \begin{tabular}[l]{@{}l@{}}\cite{mazare2018training} \end{tabular} &  \begin{minipage}[t]{\linewidth}
    \begin{itemize}[wide, labelwidth=!, labelindent=0pt]
     \item \begin{tabular}[c]{@{}l@{}}Reddit Dataset\end{tabular}
    \end{itemize}
    \end{minipage}& \begin{minipage}[t]{\linewidth}
    \begin{itemize}[wide, labelwidth=!, labelindent=0pt]
     \item \begin{tabular}[l]{@{}l@{}} Transformer \\ + Persona \\+ Context \\+ Response \\ Encoder \end{tabular}
    \end{itemize}
    \end{minipage} & \begin{minipage}[t]{\linewidth}
    \begin{itemize}[wide, labelwidth=!, labelindent=0pt]
     \item \begin{tabular}[l]{@{}l@{}} Adamax \end{tabular}
    \end{itemize}
    \end{minipage} &  \begin{minipage}[t]{\linewidth}
    \begin{itemize}[wide, labelwidth=!, labelindent=0pt]
     \item \begin{tabular}[l]{@{}l@{}} Hits@1 \end{tabular}
    \end{itemize}
    \end{minipage} \\ \midrule

 \begin{tabular}[l]{@{}l@{}} \cite{zhang2018personalizing} \end{tabular} & \begin{minipage}[t]{\linewidth}
    \begin{itemize}[wide, labelwidth=!, labelindent=0pt]
     \item \begin{tabular}[c]{@{}l@{}}PERSONA Chat \\Dataset\end{tabular}
    \end{itemize}
    \end{minipage}&  \begin{minipage}[t]{\linewidth}
    \begin{itemize}[wide, labelwidth=!, labelindent=0pt]
     \item \begin{tabular}[c]{@{}l@{}}Baseline \\Ranking \\ Models\end{tabular}
     \item \begin{tabular}[c]{@{}l@{}}Ranking \\Profile \\ Memory \\Network\end{tabular}
     \item \begin{tabular}[c]{@{}l@{}}Key-Value \\Memory \\Network\end{tabular}
     \item \begin{tabular}[c]{@{}l@{}}Seq2Seq\end{tabular}
     \item \begin{tabular}[c]{@{}l@{}}Generative \\ Profile \\ Memory \\ Network\end{tabular}
    \end{itemize}
    \end{minipage}&  \begin{tabular}[l]{@{}l@{}} N/A \end{tabular}&  \begin{minipage}[t]{\linewidth}
    \begin{itemize}[wide, labelwidth=!, labelindent=0pt]
     \item \begin{tabular}[c]{@{}l@{}}Human Evaluation\end{tabular}
     \begin{itemize}[wide, labelwidth=!, labelindent=0pt]
         \item \begin{tabular}[c]{@{}l@{}}Fluency\end{tabular}
         \item \begin{tabular}[c]{@{}l@{}}Engagingness\end{tabular}
         \item \begin{tabular}[c]{@{}l@{}}Consistency\end{tabular}
         \item \begin{tabular}[c]{@{}l@{}}Persona Detection\end{tabular}
     \end{itemize}
     \item \begin{tabular}[c]{@{}l@{}}Perplexity\end{tabular}
     \item \begin{tabular}[c]{@{}l@{}}Hits@1\end{tabular}
    \end{itemize}
    \end{minipage}\\ \midrule

 \begin{tabular}[l]{@{}l@{}}\cite{rashkin2018know} \end{tabular} &  \begin{minipage}[t]{\linewidth}
    \begin{itemize}[wide, labelwidth=!, labelindent=0pt]
     \item \begin{tabular}[l]{@{}l@{}} Empathetic Dialogues \end{tabular}
    \end{itemize}
    \end{minipage} &  \begin{minipage}[t]{\linewidth}
    \begin{itemize}[wide, labelwidth=!, labelindent=0pt]
     \item \begin{tabular}[l]{@{}l@{}} Transformer \\ Model \end{tabular}
    \end{itemize}
    \end{minipage}& \begin{minipage}[t]{\linewidth}
    \begin{itemize}[wide, labelwidth=!, labelindent=0pt]
     \item \begin{tabular}[l]{@{}l@{}} Adamax \end{tabular}
    \end{itemize}
    \end{minipage}&  \begin{minipage}[t]{\linewidth}
    \begin{itemize}[wide, labelwidth=!, labelindent=0pt]
    \item \begin{tabular}[c]{@{}l@{}}Perplexity\end{tabular}
    \item \begin{tabular}[c]{@{}l@{}}Avg BLEU\end{tabular}
    \item \begin{tabular}[c]{@{}l@{}}P@1\end{tabular}
     \item \begin{tabular}[c]{@{}l@{}}Human Evaluation\end{tabular}
     \begin{itemize}[wide, labelwidth=!, labelindent=0pt]
         \item \begin{tabular}[c]{@{}l@{}}Empathy\end{tabular}
         \item \begin{tabular}[c]{@{}l@{}}Relevance\end{tabular}
         \item \begin{tabular}[c]{@{}l@{}}Fluency\end{tabular}
     \end{itemize}
    \end{itemize}
    \end{minipage}
 \\ \midrule
 
 \begin{tabular}[l]{@{}l@{}}\cite{huang2018automatic} \end{tabular} &  \begin{minipage}[t]{\linewidth}
    \begin{itemize}[wide, labelwidth=!, labelindent=0pt]
     \item \begin{tabular}[l]{@{}l@{}} Open Subtitles \end{tabular}
     \item \begin{tabular}[l]{@{}l@{}} CBET \end{tabular}
    \end{itemize}
    \end{minipage}&  \begin{minipage}[t]{\linewidth}
    \begin{itemize}[wide, labelwidth=!, labelindent=0pt]
     \item \begin{tabular}[l]{@{}l@{}} Seq2Seq \end{tabular}
    \end{itemize}
    \end{minipage}&  \begin{minipage}[t]{\linewidth}
    \begin{itemize}[wide, labelwidth=!, labelindent=0pt]
     \item \begin{tabular}[l]{@{}l@{}} Adam \end{tabular}
    \end{itemize}
    \end{minipage} &  \begin{minipage}[t]{\linewidth}
    \begin{itemize}[wide, labelwidth=!, labelindent=0pt]
     \item \begin{tabular}[l]{@{}l@{}} Accuracy \end{tabular}
    \end{itemize}
    \end{minipage} \\ \midrule

 \begin{tabular}[l]{@{}l@{}}\cite{niu2018polite} \end{tabular} &  \begin{minipage}[t]{\linewidth}
    \begin{itemize}[wide, labelwidth=!, labelindent=0pt]
     \item \begin{tabular}[l]{@{}l@{}} Stanford Politeness \\Corpus \end{tabular}
     \item \begin{tabular}[l]{@{}l@{}} Stack Exchange \end{tabular}
    \end{itemize}
    \end{minipage}  &  \begin{minipage}[t]{\linewidth}
    \begin{itemize}[wide, labelwidth=!, labelindent=0pt]
     \item \begin{tabular}[l]{@{}l@{}}Seq2Seq \end{tabular}
     \item \begin{tabular}[l]{@{}l@{}} Fusion \\ model \\ (Seq2Seq +\\ polite-LM) \end{tabular}
     \item \begin{tabular}[l]{@{}l@{}} Label fine\\tune Model \end{tabular}
     \item \begin{tabular}[l]{@{}l@{}} Polite-RL\end{tabular}
    \end{itemize}
    \end{minipage}&  \begin{minipage}[t]{\linewidth}
    \begin{itemize}[wide, labelwidth=!, labelindent=0pt]
     \item \begin{tabular}[l]{@{}l@{}} Adam \end{tabular}
    \end{itemize}
    \end{minipage}&  \begin{minipage}[t]{\linewidth}
    \begin{itemize}[wide, labelwidth=!, labelindent=0pt]
             \item \begin{tabular}[c]{@{}l@{}}Perplexity\end{tabular}
         \item \begin{tabular}[c]{@{}l@{}}Perplexity@L\end{tabular}
         \item \begin{tabular}[c]{@{}l@{}}Word Error Rate\end{tabular}
          \item \begin{tabular}[c]{@{}l@{}}Word Error Rate@L\end{tabular}
           \item \begin{tabular}[c]{@{}l@{}}BLEU-4\end{tabular}
     \item \begin{tabular}[c]{@{}l@{}}Human Evaluation\end{tabular}
     \begin{itemize}[wide, labelwidth=!, labelindent=0pt]
         \item \begin{tabular}[c]{@{}l@{}}Politeness\end{tabular}
         \item \begin{tabular}[c]{@{}l@{}}Quality\end{tabular}
     \end{itemize}
    \end{itemize}
    \end{minipage}\\ \midrule

 \begin{tabular}[l]{@{}l@{}} \cite{chen2018hierarchical} \end{tabular} &  \begin{minipage}[t]{\linewidth}
    \begin{itemize}[wide, labelwidth=!, labelindent=0pt]
     \item \begin{tabular}[c]{@{}l@{}}Ubuntu Dialogues\end{tabular}
     \item \begin{tabular}[c]{@{}l@{}}Douban Conversation\end{tabular}
     \item \begin{tabular}[c]{@{}l@{}}JD Customer Service\end{tabular}
    \end{itemize}
    \end{minipage}&   \begin{minipage}[t]{\linewidth}
    \begin{itemize}[wide, labelwidth=!, labelindent=0pt]
     \item \begin{tabular}[c]{@{}l@{}}Hierarchical \\Variational\\ Memory \\Network\end{tabular}
    \end{itemize}
    \end{minipage}&  \begin{minipage}[t]{\linewidth}
    \begin{itemize}[wide, labelwidth=!, labelindent=0pt]
     \item \begin{tabular}[c]{@{}l@{}}Adam\end{tabular}
    \end{itemize}
    \end{minipage}&  \begin{minipage}[t]{\linewidth}
    \begin{itemize}[wide, labelwidth=!, labelindent=0pt]
     \item \begin{tabular}[c]{@{}l@{}}Human Evaluation\end{tabular}
     \begin{itemize}[wide, labelwidth=!, labelindent=0pt]
         \item \begin{tabular}[c]{@{}l@{}}Appropriateness\end{tabular}
         \item \begin{tabular}[c]{@{}l@{}}Informativeness\end{tabular}
     \end{itemize}
          \item \begin{tabular}[c]{@{}l@{}}Embedding Evaluation\end{tabular}
     \begin{itemize}[wide, labelwidth=!, labelindent=0pt]
         \item \begin{tabular}[c]{@{}l@{}}Average\end{tabular}
         \item \begin{tabular}[c]{@{}l@{}}Greedy\end{tabular}
          \item \begin{tabular}[c]{@{}l@{}}Extrema\end{tabular}
     \end{itemize}
    \end{itemize}
    \end{minipage} \\ \midrule

  \begin{tabular}[l]{@{}l@{}}\cite{ghazvininejad2018knowledge} \end{tabular} &   \begin{minipage}[t]{\linewidth}
    \begin{itemize}[wide, labelwidth=!, labelindent=0pt]
     \item \begin{tabular}[c]{@{}l@{}}Twitter Conversation \\Dialogues\end{tabular}
     \item \begin{tabular}[c]{@{}l@{}}Four Square\end{tabular}
    \end{itemize}
    \end{minipage} &  \begin{minipage}[t]{\linewidth}
    \begin{itemize}[wide, labelwidth=!, labelindent=0pt]
     \item \begin{tabular}[c]{@{}l@{}}Seq2Seq + \\ World Facts \\+ Contextual \\ Facts \end{tabular}
    \end{itemize}
    \end{minipage}& \begin{minipage}[t]{\linewidth}
    \begin{itemize}[wide, labelwidth=!, labelindent=0pt]
     \item \begin{tabular}[c]{@{}l@{}}Adam\end{tabular}
    \end{itemize}
    \end{minipage} &  \begin{minipage}[t]{\linewidth}
    \begin{itemize}[wide, labelwidth=!, labelindent=0pt]
             \item \begin{tabular}[c]{@{}l@{}}Perplexity\end{tabular}
         \item \begin{tabular}[c]{@{}l@{}}BLEU\end{tabular}
         \item \begin{tabular}[c]{@{}l@{}}Diversity\end{tabular}
     \item \begin{tabular}[c]{@{}l@{}}Human Evaluation\end{tabular}
     \begin{itemize}[wide, labelwidth=!, labelindent=0pt]
         \item \begin{tabular}[c]{@{}l@{}}Informativeness\end{tabular}
         \item \begin{tabular}[c]{@{}l@{}}Appropriateness\end{tabular}
     \end{itemize}
    \end{itemize}
    \end{minipage} \\ \midrule
 
  \begin{tabular}[l]{@{}l@{}} \cite{young2018augmenting} \end{tabular} &  \begin{minipage}[t]{\linewidth}
    \begin{itemize}[wide, labelwidth=!, labelindent=0pt]
     \item \begin{tabular}[c]{@{}l@{}}Twitter Conversation \\ Dialogues\end{tabular}
    \end{itemize}
    \end{minipage} &  \begin{minipage}[t]{\linewidth}
    \begin{itemize}[wide, labelwidth=!, labelindent=0pt]
     \item \begin{tabular}[c]{@{}l@{}}Tri-LSTM \\ Encoder\end{tabular}
    \end{itemize}
    \end{minipage} &  \begin{minipage}[t]{\linewidth}
    \begin{itemize}[wide, labelwidth=!, labelindent=0pt]
     \item \begin{tabular}[c]{@{}l@{}}SGD\end{tabular}
    \end{itemize}
    \end{minipage}&  \begin{minipage}[t]{\linewidth}
    \begin{itemize}[wide, labelwidth=!, labelindent=0pt]
     \item \begin{tabular}[c]{@{}l@{}}Recall@k\end{tabular}
    \end{itemize}
    \end{minipage} \\ \midrule
   
   \begin{tabular}[l]{@{}l@{}}\cite{dinan2018wizard} \end{tabular} &  \begin{minipage}[t]{\linewidth}
    \begin{itemize}[wide, labelwidth=!, labelindent=0pt]
     \item \begin{tabular}[c]{@{}l@{}}Wizards of Wikipedia\end{tabular}
    \end{itemize}
    \end{minipage} & \begin{minipage}[t]{\linewidth}
    \begin{itemize}[wide, labelwidth=!, labelindent=0pt]
     \item \begin{tabular}[c]{@{}l@{}}Retrieval\\ Transformer \\ Memory \\Network\end{tabular}
     \item \begin{tabular}[c]{@{}l@{}}Generative \\Transformer \\ Memory \\Network\end{tabular}
    \end{itemize}
    \end{minipage} & \begin{minipage}[t]{\linewidth}
    \begin{itemize}[wide, labelwidth=!, labelindent=0pt]
     \item \begin{tabular}[c]{@{}l@{}}NLL\end{tabular}
    \end{itemize}
    \end{minipage} &  \begin{minipage}[t]{\linewidth}
    \begin{itemize}[wide, labelwidth=!, labelindent=0pt]
             \item \begin{tabular}[c]{@{}l@{}}Recall@1\end{tabular}
         \item \begin{tabular}[c]{@{}l@{}}Perplexity\end{tabular}
     \item \begin{tabular}[c]{@{}l@{}}Human Evaluation\end{tabular}
     \begin{itemize}[wide, labelwidth=!, labelindent=0pt]
         \item \begin{tabular}[c]{@{}l@{}}Engagingness\end{tabular}
     \end{itemize}
    \end{itemize}
    \end{minipage}
\\ \midrule

 \begin{tabular}[l]{@{}l@{}}\cite{wolf2019transfertransfo} \end{tabular} &  \begin{minipage}[t]{\linewidth}
    \begin{itemize}[wide, labelwidth=!, labelindent=0pt]
     \item \begin{tabular}[c]{@{}l@{}}PERSONA Chat \\ Dataset \end{tabular}
    \end{itemize}
    \end{minipage} &  \begin{minipage}[t]{\linewidth}
    \begin{itemize}[wide, labelwidth=!, labelindent=0pt]
     \item \begin{tabular}[c]{@{}l@{}}Transformer \\ Model\end{tabular}
    \end{itemize}
    \end{minipage}& \begin{minipage}[t]{\linewidth}
    \begin{itemize}[wide, labelwidth=!, labelindent=0pt]
     \item \begin{tabular}[c]{@{}l@{}}Adam\end{tabular}
    \end{itemize}
    \end{minipage} &  \begin{minipage}[t]{\linewidth}
    \begin{itemize}[wide, labelwidth=!, labelindent=0pt]
     \item \begin{tabular}[c]{@{}l@{}}Perplexity\end{tabular}
     \item \begin{tabular}[c]{@{}l@{}}Hits@1\end{tabular}
     \item \begin{tabular}[c]{@{}l@{}}F1 Score\end{tabular}
    \end{itemize}
    \end{minipage}  \\ \midrule

 \begin{tabular}[l]{@{}l@{}}\cite{zheng2019personalized} \end{tabular} &  \begin{minipage}[t]{\linewidth}
    \begin{itemize}[wide, labelwidth=!, labelindent=0pt]
     \item \begin{tabular}[c]{@{}l@{}}PERSONALDIALOG\\ Dataset\end{tabular}
    \end{itemize}
    \end{minipage}& \begin{minipage}[t]{\linewidth}
    \begin{itemize}[wide, labelwidth=!, labelindent=0pt]
     \item \begin{tabular}[c]{@{}l@{}}Seq2Seq + \\ Personality \\Fusion \end{tabular}
    \end{itemize}
    \end{minipage} &  \begin{minipage}[t]{\linewidth}
    \begin{itemize}[wide, labelwidth=!, labelindent=0pt]
     \item \begin{tabular}[c]{@{}l@{}}Adam \end{tabular}
    \end{itemize}
    \end{minipage} &  \begin{minipage}[t]{\linewidth}
    \begin{itemize}[wide, labelwidth=!, labelindent=0pt]
             \item \begin{tabular}[c]{@{}l@{}}Perplexity\end{tabular}
         \item \begin{tabular}[c]{@{}l@{}}Distinct-1\end{tabular}
         \item \begin{tabular}[c]{@{}l@{}}Distinct-2\end{tabular}
         \item \begin{tabular}[c]{@{}l@{}}Accuracy\end{tabular}
    \item \begin{tabular}[c]{@{}l@{}}Human Evaluation\end{tabular}
     \begin{itemize}[wide, labelwidth=!, labelindent=0pt]
         \item \begin{tabular}[c]{@{}l@{}}Fluency\end{tabular}
         \item \begin{tabular}[c]{@{}l@{}}Appropriateness\end{tabular}
     \end{itemize}
    \end{itemize}
    \end{minipage}\\ \bottomrule

\end{longtable}
\end{footnotesize}

\section{Conclusion}
\label{sumftd}

In this work, we summarized the work done in the area of language generation, starting from traditional approaches through recent work using deep learning approaches. Even with the rapid advancement in this sub-field of natural language generation for open domain dialogue systems, many of the approaches were based on the historical findings and prior research. We provided a summary of the important contributions by the standard Reiter and Dale \citep{reiter2000building} architecture and provided explanations and body of research conducted to address the six different components of the architecture.

Another important aspect of this summarized work is identifying potential research gaps that persist in the field of conversational agents. We propose that tackling them would advance the field further.

\subsection{Open Challenges for Open-Domain Dialogue Systems}
Even though prior work done in the area of the open domain dialogue systems have helped advance the field, there are specific issues that affect their quality. We identify three main issues:
\begin{enumerate}
    \item \textbf{Encoding Context} - Encoding contextual information such as world facts from knowledge bases or previous turns of the conversation are important issues to ensure that the conversational agent has enough information to produce a coherent, informative and novel response that is in tune with the context of the conversation. From Table 1, we find that a lot of prior research used a one-to-one mapping between a single input utterance and the generated response. This makes it hard to judge the quality of the response generated or the performance of the model with regards to the context of the conversation or how the model would perform when it comes to multi-turn conversations. 
    
    To overcome this issue, researchers have focused on including the previous turns of the conversation as contextual information to the model. This has been accomplished in two different ways: through sequential models  \citep{sordoni2015neural} and through hierarchical models  \citep{serban2016building}. In sequential encoding of the context, the previous turn of the conversation is concatenated to the current input utterance. In hierarchical encoding of the context, a two-step approach is followed by performing an utterance-level encoding followed by an inter-utterance encoding. Tian \emph{et al.}, \citeyearpar{tian2017make} conducted an empirical study that evaluated the advantages and disadvantages of sequential and hierarchical models and show that the hierarchical models outperform sequential models when encoding contextual information. 
    
    Encoding factual knowledge to augment the model was demonstrated by Dinan \emph{et al.}  \citeyearpar{dinan2018wizard} and Young \emph{et al.} \citeyearpar{young2018augmenting} using the transformer models and Tri-LSTM encoder approach respectively.
    
    \item \textbf{Incorporating Personality} - Endowing conversational agents with a coherent persona is key to building a engaging and convincing conversational agent \citep{niu2018polite}. The concept of personality has been well studied in the psychology. Traditionally, research on using personality traits has been based on the standard \textbf{\textit{Big Five}} model (extraversion, neuroticism, agreeableness, conscientiousness, and openness to experience) and some of the early works on building personalized dialogue systems have been based on the Big Five model  \citep{mairesse2007personage}. 
    
    However, identifying personality traits through \textbf{\textit{Big Five}} model is difficult and expensive to obtain \citep{zheng2019personalized,zhang2018personalizing}. Alternative approaches that take advantage of the psycholinguistics are still in their infancy. Some of the proposed approaches to solve this problem have been through explicit or implicit modeling of personality \citep{zheng2019personalized}. Explicit modeling involves creating profiles of users with features such as age, gender \citep{zheng2019personalized} or assigning artificial persona to users and asking them to interact  \citep{zhang2018personalizing}. Implicit Modeling of persona involves creating vectors about the users based on similar features such as age, gender and other personal information \citep{li2016persona,kottur2017exploring}. 
    
     More recently, the transformer models have been used for conversational agents  \citep{wolf2019transfertransfo,dinan2018wizard,rashkin2018know}. Wolf \emph{et al.} \citeyearpar{wolf2019transfertransfo} demonstrated the usage of transformer model for personalized response generation on the PERSONA-CHAT dataset where the model concatenates each artificial persona provided along with the utterances of the conversation. 
     
    \item \textbf{Dull and generic responses} - One problem with building end-to-end conversational agents based on vanilla seq2seq is that they are prone to generating dull and generic responses such as {\footnotesize{\fontfamily{pcr}\selectfont{I don't know}}}, {\footnotesize{\fontfamily{pcr}\selectfont{I am not sure.}}} etc.  \citep{vinyals2015neural,li2015diversity}. These trivial responses make the conversational agents unable to sustain longer conversations with a human. Li \emph{et al.} \citeyearpar{li2015diversity} suggested a mechanism to overcome this issue with an optimization function (see equation \ref{MLE}, where $T$ is target and $S$ is source sentence). The authors only considered likelihood of the responses when given an input and proposed using Maximum Mutual Information (MMI) as the optimization objective function (see equation \ref{MMI}) where $\lambda$ is a hyperparameter to penalize generic responses .
    \begin{equation}
    \hat{T} = \arg\max_T \{log p(T|S)\}
    \label{MLE}
    \end{equation}
    
    \begin{equation}
        \hat{T} = \arg\max_T \{log p(T|S) - \lambda log p(T) \}
        \label{MMI}
    \end{equation}
    
    Recent approaches toward conversational modeling have all tackled the issue of dull and generic response through the use of previous utterance as contextual information or with the help of attention mechanism that focuses on a particular part of the input utterance or using reinforcement learning that penalizes the agent when it produces trivial or repetitive utterance \citep{li2016deep,li2017adversarial,liu2018towards}.
\end{enumerate}

\subsection{Future Directions}
Having identified three open challenges in the subsection above, we now propose two promising future directions on how to tackle these open challenges. 
\begin{enumerate}
    \item \textbf{Cognitive Architectures} \textendash We argue that natural language generation entails not only incorporating fundamental aspects of \textit{artificial intelligence} but also \textit{cognitive science} \citep{reiter2000building, sun2007importance}. The role of cognitive architectures(CA) that offers a different perspective has not been explored for deep learning architectures. Cognitive architectures provides a blueprint for building intelligent agents by modelling human behavior. One prominent model in cognitive architecture is the Standard Model (Figure \ref{fig:standard}) \citep{norris2017short,laird2017standard}. This model provides the framework with which to conceptually and practically address both long-term memory and short-term memory (also known as working memory), along with an action-selection mechanism acting as a bridge between them. According to this model of human cognition, given an input (for example, through perception), an output is generated by taking into account elements stored in the working memory as well as long-term storage. 
    
\begin{figure*}[t]
  \centering
   \includegraphics[height=10cm,width=10cm,keepaspectratio]{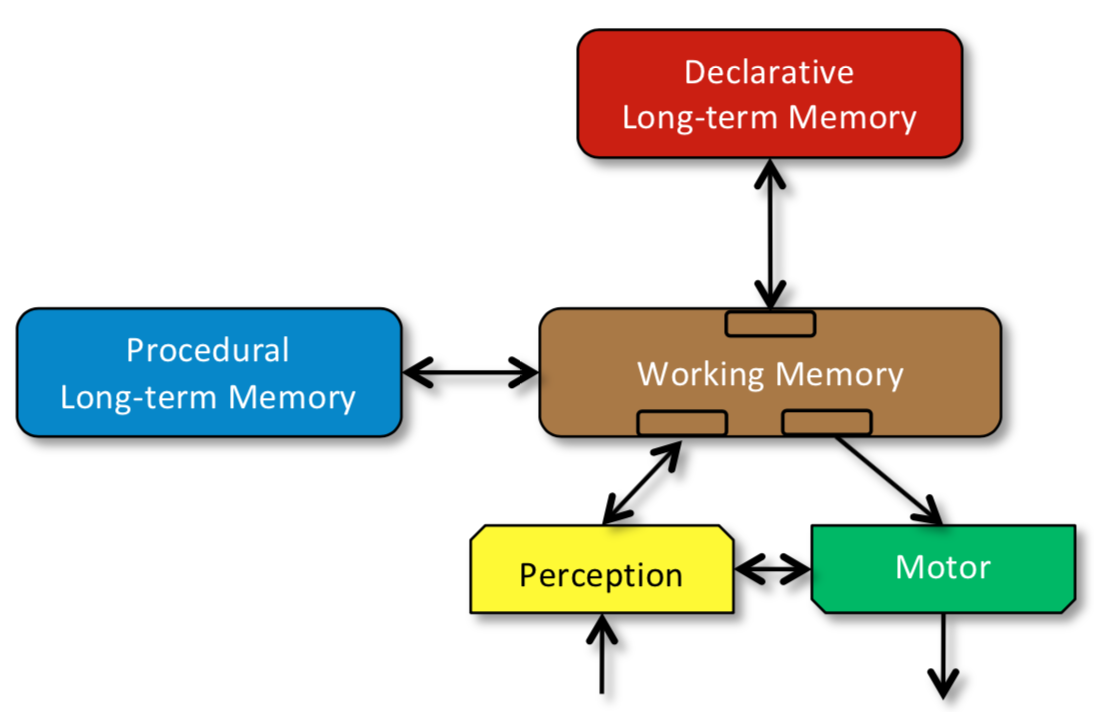}
  \caption{Standard Model of Cognitive Architecture containing two forms of long-term memory (Procedural and Declarative) and Working Memory to address given input. Figure credit: Laird \emph{et al.} \citeyearpar{laird2017standard}}
  \label{fig:standard}
\end{figure*}
    
    \item \textbf{Encoding Emotional Content} \textendash Emotions are recognized as functional in decision-making by influencing motivation and action selection \citep{Moerland2018}. Therefore, computational emotion models should be grounded in the agent`s decision making architecture.  For example, Badoy \emph{et al.}, \citeyearpar{Badoy2014QLearningWB} proposed using four basic emotions: joy, sadness, fear, and anger to influence a Qlearning agent. Simulations show that the proposed affective agent required fewer steps to find the optimal path.  In language generation work, Zhuo \emph{et al.}, \citeyearpar{zhou2018emotional} have proposed Emotional Chatting Machine (ECM) that can generate appropriate responses not only in content (relevant and grammatical) but also in emotion (emotionally consistent).  ECM addresses the factor using three new mechanisms that respectively (1) models the high-level abstraction of emotion expressions by embedding emotion categories, (2) captures the change of implicit internal emotion states, and (3) uses explicit emotion expressions with an external emotion vocabulary. Experiments show that the proposed model can generate responses appropriate not only in content but also in emotion. However, the problem of generating emotionally appropriate responses in longer conversations is still to be explored.
\end{enumerate}

We hypothesize that by incorporating elements of cognitive architectures and adding emotional content, researchers can address the open challenges that we have identified in the prior section. By adapting deep learning approaches to closely mirror the the memory mechanisms as postulated by the Standard Model (Figure~\ref{fig:standard}), dialogue systems can take advantage of longer conversational context as well as world and domain knowledge from databases. By incorporating emotional content, the challenge of having conversational agents mirror a personality can be addressed. In future work, we aim to address the open challenges through these two directions. 

\section*{Acknowledgements}
This work was supported by the Defense Advanced Research Projects Agency (DARPA) under Contract No FA8650-18-C-7881. All statements of fact, opinion or conclusions contained herein are those of the authors and should not be construed as representing the official views or policies of AFRL, DARPA, or the U.S. Government.
\maketitle





\bibliography{dnd-article-template}

\end{document}